\pdfoutput=1
\documentclass[11pt]{article}

\usepackage{acl}

\usepackage{times}
\usepackage{latexsym}

\usepackage[T1]{fontenc}

\usepackage[utf8]{inputenc}

\usepackage{microtype}

\usepackage{float}
\usepackage{inconsolata}
\usepackage{amsmath}
\usepackage{amssymb}
\usepackage{enumitem}
\usepackage{hyperref}
\newtheorem{proposition}{Proposition}

\usepackage{graphicx}
\usepackage{bbold}
\usepackage{multicol}
\usepackage{comment}
\usepackage{multirow}
\usepackage{colortbl}
\usepackage{makecell}
\usepackage{tikz}
\usetikzlibrary{shapes,arrows,positioning}

\usepackage{color}
\newcommand\mat[1]{{\textcolor{black}{#1}}}
\newcommand\rom[1]{{\textcolor{black}{#1}}}

\newcommand\roma[1]{{\textcolor{black}{#1}}}
\newcommand\matt[1]{{\textcolor{black}{#1}}}

\newcommand\romano[1]{{\textcolor{black}{#1}}}

\title{Revisiting Hierarchical Text Classification: Inference and Metrics}

\author{
    Roman Plaud\textsuperscript{1,2}, Matthieu Labeau\textsuperscript{1}, Antoine Saillenfest\textsuperscript{2}, Thomas Bonald\textsuperscript{1} \\
    \textsuperscript{1} Institut Polytechnique de Paris \\ 
    \textsuperscript{2} Onepoint, 29 rue des Sablons, 75016, Paris, France \\
    \texttt{\{roman.plaud, matthieu.labeau, thomas.bonald\}@telecom-paris.fr} \\     \texttt{a.saillenfest@groupeonepoint.com}
}

\begin{document}
\maketitle
\begin{abstract}
Hierarchical text classification (HTC) is the task of assigning labels to a text within a structured space organized as a hierarchy. Recent works
treat HTC as a conventional multilabel classification problem, therefore evaluating it as such.
We instead propose to evaluate models based on specifically designed hierarchical metrics and we demonstrate the intricacy of metric choice and prediction inference method.  
We introduce a new challenging dataset and we 
evaluate recent sophisticated models against a range of simple but strong baselines, including a new theoretically motivated loss. 
Finally, we show that those baselines 
are very often competitive with the latest models. This highlights the importance of carefully considering the evaluation methodology 
when proposing new methods for HTC. Code implementation and dataset are available at \url{https://github.com/RomanPlaud/revisitingHTC}.
\end{abstract}
\section{Introduction}
 Text classification is a long-studied problem that \mat{may involve various types of label sets.}
In particular, Hierarchical Text Classification (HTC) includes labels that exhibit a hierarchical structure with parent-child relationships. The structure that emerges from these relationships is either a tree \cite{WOS, rcv1, bugs, bgc, nyt} or a Directed Acyclic Graph (DAG) \cite{bertinetto2020}. Each input text then comes with a set of labels that form one or more paths in the hierarchy. 
A first crucial challenge in HTC lies in accurately evaluating model performance. This requires metrics that are sensitive to the severity of prediction errors, penalizing mistakes with larger distances within the hierarchy tree. While pioneering efforts have been made by \citet{hierf1score}, \citet{silla_freitas}, \citet{evaluation_metrics} and \citet{amigo-delgado-2022-evaluating}, evaluation in the context of hierarchical classification remains an ongoing research area.
\begin{figure}[!t]
    \centering
    \includegraphics[width=\columnwidth]{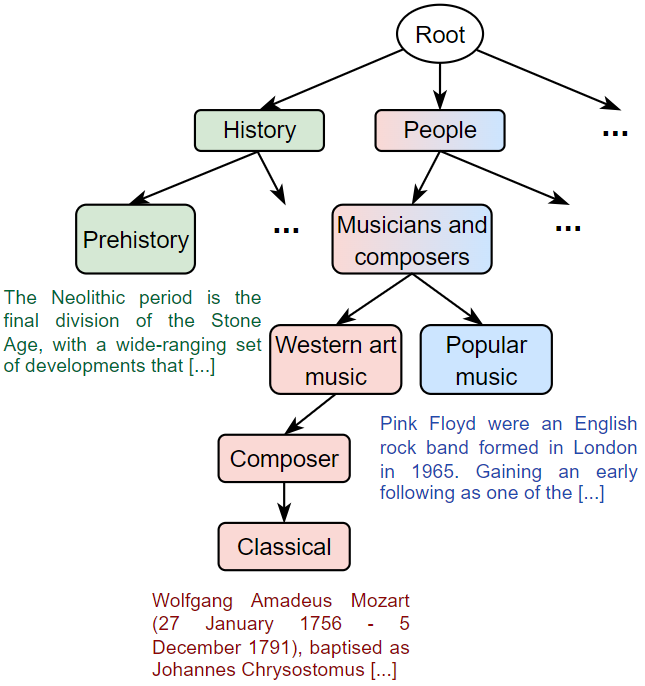}
    \caption{Extract of the taxonomy of our new dataset Hierarchical WikiVitals. Each colored path is the set of labels of the same color.}
    \label{fig:enter-label}
\end{figure} There is a substantial body of literature addressing HTC. \mat{The most recent methods produce text representations which are \textit{hierarchy-aware}, as they integrate information about the label hierarchy~\cite{song-etal-2023-peer, zhou-etal-2020-hierarchy, deng2021htcinfomax, wang-etal-2022-hpt, wang2022incorporating, jiang-etal-2022-exploiting, HiMatch, zhu2023hitin, zhu-etal-2024-hill, yu-etal-2023-instances}. However, we believe that the evaluation of these models has been insufficiently investigated: \matt{in those works, the task is evaluated as standard multi-label classification.}}
Here, we plan to \matt{explore what this implies; especially, looking at how predictions are inferred from an estimated probability distribution} -- which we consider an under-addressed challenge. 
We provide new insights, emphasizing the intricacy of inference and evaluation, which cannot be considered separately. To complete this investigation, we introduce a new English benchmark dataset, Hierarchical WikiVitals (HWV), which \mat{we intend to} be significantly more challenging than 
the usual HTC benchmarks in English (see Figure~\ref{fig:enter-label} for an extract of the taxonomy).
\mat{We experiment within our proposed framework, verifying the performance of recent models against simpler methods, 
among which loss functions~\cite{bertinetto2020, vaswani2022champ, zhang2021match} we design to be able to integrate hierarchical information, based on the conditional softmax.} 
Overall, our contributions are:

\roma{\begin{enumerate}[topsep=-\parskip, leftmargin=*]
    \itemsep0em 
    \item We propose to quantitatively evaluate HTC methods based on specifically designed hierarchical metrics and with a rigorous methodology.
    \item We present Hierarchical WikiVitals, a novel high-quality HTC dataset, extracted from Wikipedia. Equipped with a deep and complex hierarchy, it provides a harder challenge.
    \item We conduct extensive experiments on three popular HTC datasets and HWV, introducing a novel loss function. When combined with a BERT model, this approach achieves competitive results against recent advanced models.
\end{enumerate}}
\noindent\matt{Our results show that state-of-the-art models do not necessarily encode hierarchical information well, and are surpassed by our simpler loss on HWV.}

\subsection*{Problem definition}
\label{sec:prob_def}
Hierarchical Text classification (HTC) is a subtask of text classification which consists of assigning to an input text $x \in \mathcal{X}$ a set of labels $Y \subset \mathcal{Y}$, where the label space $\mathcal{Y}$ exhibits parent-child relationships. We call hierarchy the directed graph $\mathcal{H} = (\mathcal{Y}, \mathcal{E})$, where  $\mathcal{E} \subset \mathcal{Y}^2$ is the set of edges, which goes from a parent to its children. We restrain our study to the case where $\mathcal{H}$ is a tree. We follow the notations of \citet{NEURIPS2022_727855c3} and call $\mathbf{r} \in \mathcal{Y}$ the unique root node and $\mathcal{L}$ the set of leaf nodes. For a node $y \in \mathcal{Y}\backslash\{\mathbf{r}\}$ we denote $\pi(y)$ its unique parent, $\mathcal{C}(y) \subset \mathcal{Y}$ the set of its children and $\mathcal{A}(y)$ the set of its ancestors (defined inclusively). 

 A label set $Y$ of an input $x$ cannot be arbitrary: if $y\in Y$ then, due to the parent relations, we necessarily observe that $\mathcal{A}(y)\subset Y$. 
An even more restrictive framework is the \textit{single-path leaf labels} setting, where  $Y=\mathcal{A}(l)$ for a given $l\in\mathcal{L}$ ($Y$ is a single path and reaches a leaf). 

 We study methods mapping an input text $x$ to a conditional distribution $\mathbb{P}(\cdot|x)$ over $\mathcal{Y}$, whose estimation is denoted $\hat{\mathrm{P}}(\cdot|x)$.
\matt{Lastly, what we call \textit{inference rule} is the way of producing a set of binary predictions from   a probability distribution. For example predictions can be obtained by thresholding $\hat{\mathrm{P}}(\cdot|x)$ to $\tau$ as follows :  $\hat{Y}_{\tau} = \{ y \in \mathcal{Y},  \hat{\mathrm{P}}(y|x) > \tau \}$.} 
\section{Related Work}
\subsection{Hierarchical Text Classification}
Hierarchical classification problems, including the particular case of HTC, are typically dealt with through either a \textit{local} approach or a \textit{global} one. We refer to the original definition made by \citet{silla_freitas} according to which the difference between the two categories lies in the training phase. Indeed, local methods imply training a collection of specialized classifiers, \textit{e.g.} one for each node, for each parent node or even one for each level; and during its training each classifier is unaware of the holistic structure of the hierarchy~\cite{review}. While often computationally costly, it has proven to be effective to capture crucial local information. Along those lines,~\citet{banerjee-etal-2019-hierarchical} propose to link the parameters of a parent classifier and those of its children, following the idea of transferring knowledge from parent nodes to their descendants~\cite{shimura-etal-2018-hft, huang, pmlr-v80-wehrmann18a}. 
Conversely, global methods involve a unique model that directly incorporates the whole hierarchical information in their predictions. 
There exist very different types of global approaches, from which we can draw two broad categories: losses incorporating hierarchical penalties and hierarchy-aware models.
\\
\textbf{Hierarchical \mat{penalties}}.
The idea of these methods is generally to use a standard binary cross-entropy (BCE), and add penalization terms that incorporate hierarchical information. \citet{recursive_reguluation_Gopal_2013} and \citet{zhang2021match} propose regularization based on hypernymy, either acting on the parameter space or the outputted probability space, while \citet{vaswani2022champ} introduce an enhanced BCE loss, named CHAMP, which penalizes false positives based on their distance to the ground truth in the hierarchy tree. \\
\textbf{Hierarchy-aware models}. 
To incorporate the structural constraints of the hierarchy into prediction, \citet{mao-etal-2019-hierarchical} propose a reinforcement learning approach, while \citet{bgc} introduce an architecture based on capsule networks. However, recent works have achieved state-of-the-art results by combining a text encoder with a structure encoder applied to the label hierarchy. This concept was first introduced by \citet{zhou-etal-2020-hierarchy}, who utilized graph convolution networks as the hierarchy encoder. Building on this foundational work, \citet{jiang-etal-2022-exploiting} and \citet{wang2024utilizing} developed methods to better incorporate local hierarchy information. \citet{wang2022incorporating} proposed a contrastive learning approach, while \citet{zhu2023hitin} designed a method to encode hierarchy with the guidance of structural entropy. \citet{zhu-etal-2024-hill} combined both of these ideas. These developments follow earlier works on the same concept \cite{chen2019hyperbolic, ZHANG2022115922, deng2021htcinfomax, HiMatch, WANG2021106876}. It is important to note that these models are typically trained with a BCE loss or one of its penalized versions \cite{zhang2021match}.

\subsection{Hierarchical prediction}
Making a prediction in HTC involves two
seemingly irreconcilable difficulties: 
one has to decide between making independent predictions, which may lead to \textit{coherence} issues (e.g., predicting a child without predicting its parent), or employing a top-down inference approach, which may cause \textit{error propagation} issues~\cite{yang-cardie-2013-joint, song-etal-2012-cost}. 
Recent hierarchy-aware models predominantly operate within the former framework, 
training and evaluating the model as a simple multi-label classifier, at the price of ignoring potentially badly structured predictions.
In this work, we will experiment with both approaches. 
\subsection{Hierarchical classification evaluation}
\roma{Evaluation in the context of hierarchical classification is a long-studied problem \cite{evaluation_metrics, amigo-delgado-2022-evaluating, review_metrics} from which arise multiple questions. First, diverse setups exist, \matt{implying different assumptions on the labeling structure: while} 
we \matt{previously} introduced the \textit{single-path leaf label} framework, multi-path hierarchies exist, or even inputs with only non-leaf labels. It is therefore important to design metrics that are \textbf{agnostic to the hierarchical classification framework}. Then, a hierarchical metric must indeed be hierarchical. This means it should take into account the severity of an error based on the known hierarchy: intuitively, predicting a \textit{Bulldog} instead of a \textit{Terrier} should be less penalized than predicting a \textit{Unicorn} instead of a \textit{Terrier}. \citet{amigo-delgado-2022-evaluating} identify a set of properties an evaluation metric should possess for hierarchical classification, and classifies them in a taxonomy of metrics differentiating between \textbf{multi-label metrics} (label-based, example-based, ranking-metrics) and \textbf{hierarchical metrics} (pair-based, set-based). We heavily rely on this seminal work when it comes to choose which metric to use to evaluate different methods. Finally, the inference rule 
\matt{should be chosen in accordance with the metric.} 
The bayesian decision theory literature \cite{berger1985statistical} aims at finding an optimal 
rule given the metric of interest. However, little consideration was given to this issue in the context of hierarchical classification and ad hoc and non-statistically grounded inference methodology are often chosen: for example, recent HTC literature mostly performs inference through thresholding the estimated probability distribution \matt{with $\tau = 0.5$}. 
We can think of other inference methodology, based on top-down or bottom-up inference rules. It is then crucial to find metrics that either come with a properly grounded prediction rule, or \textbf{do not depend on an inference methodology} but rather account for the whole probability distribution, 
\matt{which implies} evaluating at different operating points.
In the next part, we will re-introduce metrics in the light of 
the three listed requirements.  }
\section{Evaluation metrics}

\roma{\matt{The aforementioned inference rule used in recent HTC literature corresponds to a classical multi-label evaluation methodology: computing a F1-score (\textit{micro} and \textit{macro}) with $\tau = 0.5$. 
In what follows, we show that this thresholding scheme is 
suboptimal and we introduce the metrics we use in our experiments. 
We will then motivate the use of an inference-free evaluation methodology.}} 

\subsection{Multi-label metrics}
There is a large array of methods for multi-label evaluation;~\citet{Wu2016AUV}, through unifying notations, proposed a set of 11 different metrics. Among them, we keep the \textit{micro} and \textit{macro} F1-score \mat{computed upon scores obtained through a  $0.5$ threshold}, as it is generally done in HTC literature.
We add a simple metric corresponding to the fraction of misclassified labels: the Hamming Loss, which we also couple to a $0.5$ thresholding inference rule.\footnote{This optimal inference holds in case of label independence \cite{hammingloss} which is not the case here.} 

\subsection{Hierarchical metrics}
\label{sec:hier_f1}
We introduce hF1-score which we identify to be relevant to our evaluation framework.  
We note that 
a prediction is \textit{coherent} if $z\in \hat{Y} \Rightarrow \mathcal{A}(z) \subset \hat{Y}$. 

\paragraph{Hierarchical F1-score.} Introduced by \citet{hierf1score}, this \textbf{set-based} measure consists in augmenting $\hat{Y}$ with all its ancestors as follows : 
\begin{equation}
    \hat{Y}^{\text{aug}} = \underset{\hat{y}\in\hat{Y}}{\cup}\mathcal{A}(\hat{y})
\end{equation}
And to compute the hierarchical precision, recall and F1-score are as follows : 
\begin{equation*}
        \mathrm{hP(Y, \hat{Y})} = \frac{\left|\hat{Y}^{\text{aug}}\cap Y\right|}{\left|\hat{Y}^{\text{aug}}\right|} ~ \mathrm{hR(Y, \hat{Y})}=\frac{\left|\hat{Y}^{\text{aug}}\cap Y\right|}{\left|Y\right|}
\end{equation*}
\begin{equation*}
    \mathrm{hF1(Y, \hat{Y})} = \frac{2\cdot\mathrm{hP(Y, \hat{Y})} \cdot \mathrm{hR(Y, \hat{Y})}}{\mathrm{hP(Y, \hat{Y})} + \mathrm{hR(Y, \hat{Y})}}
\end{equation*}
 It is a simple extension of the F1-score to hierarchical classification. In the multi-label setting, there are several methods of aggregation to compute a global F1-score\footnote{See for example the \href{https://scikit-learn.org/stable/modules/generated/sklearn.metrics.f1_score.html}{Scikit-learn documentation}.}. We define here a per-instance hF1-score as per \citet{evaluation_metrics} which is then averaged over all inputs (referred as \textit{samples} setting). In its very first introduction, it was defined in a \textit{micro} fashion by ~\citet{hierf1score} (see Appendix~\ref{app:equivalence} for full definitions).

\begin{proposition}
\label{prop:1}
    In \textit{micro} and \textit{samples} settings, if every prediction $\hat{Y}$ is coherent, then hF1 and F1 are strictly equal.
\end{proposition}
\textbf{Motivations}. Hierarchical F1-score considers an ancestor overlap between ground truth and predicted labels therefore accounting for \textbf{mistake severity} and is also \textbf{agnostic to the hierarchical classification framework}. Moreover, Proposition~\ref{prop:1} (whose proof is detailed in Appendix~\ref{app:equivalence}) draws a link between example-based multi-label metrics and set-based hierarchical metrics proving that
it was therefore relevant to employ the \textit{micro} F1-score as it is done in recent literature, {as long as predictions are coherent}. Finally, hF1-score incorporates \textbf{all desirable hierarchical properties} as listed by~\citet{amigo-delgado-2022-evaluating}, except that it does not completely capture the \textit{specificity} (\textit{i.e} the level of uncertainty left by predicting a given node). 

\paragraph{Other hierarchical metrics.} As explained in previous section, hF1-score is imperfect as it assumes an equivalence between depth and specificity. To solve this issue,~\citet{NEURIPS2022_727855c3} has proposed an information-based hierarchical F1-score, introduced in Appendix~\ref{app:hier_metrics_not_used}. 
\romano{There also exist constrained versions of \textbf{multi-label F1-scores} \citep{C_F1_v2, C_F1} which account for coherence issues: a correct prediction for a label node is valid only if all its ancestor nodes are correct predictions.\\ Although these metrics might seem pertinent, we have chosen not to utilize them, as they do not globally influence the ranking of methods when compared to their standard metric counterparts. We thoroughly detail our reasons in Appendix~\ref{app:hier_metrics_not_used}. An important number of context-dependent hierarchical metrics were also introduced~\citep{sunlin2001, 7118216}, which we will not discuss here as we aim for agnosticism to the hierarchical classification~context.}

\subsection{Inference methodology}

\begin{figure}
    \centering
    \includegraphics[width=\columnwidth]{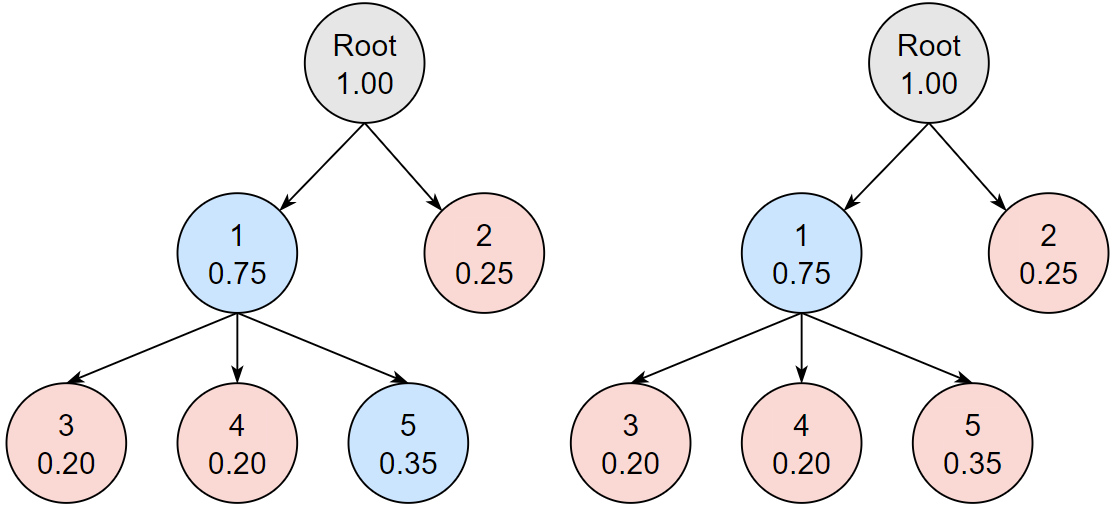}
    \caption{Example of a conditional distribution estimation over a simple hierarchy and corresponding predicted nodes (in blue) for different thresholds ($0.3$ \mat{on the left}, $0.5$ \mat{on the right}).}
    \label{fig:ex_graph}
\end{figure}

In this section, we begin by \matt{motivating our argument} against the practice of using a BCE-based loss and 
 $\tau = 0.5$ to produce predictions. While this corresponds to minimizing the multilabel Hamming loss in case of label independence~\citep{hammingloss}, 
there is to the best of our knowledge no evidence of the optimality of such a predictor in a hierarchical setting. \matt{Rather, tools such as \textit{risk minimization} can provide a way to obtain a statistically grounded inference methodology optimizing the chosen metric, from an estimation of $\mathbb{P}(\cdot|x)$, obtained by a model for a given $x$. In particular, it is possible to show that the optimal threshold for the F1-scores depends on $\mathbb{P}(\cdot|x)$; we detail the proof in Appendix~\ref{sec:dep_x}. Though, a simple counter-example is enough to invalidate the choice of $0.5$: such an example is depicted in Figure~\ref{fig:ex_graph}. It shows a coherent and exhaustive probability distribution $\mathbb{P}(\cdot|x)$, for a given $x$. Thresholding to $0.5$ would lead to predict $\{1\}$, while a simple computation, detailed in Appendix~\ref{sec:inf_methodo}, gives:
\begin{align*}
&\mathbb{E}[\mathrm{hF1(Y}, \{1\})|X=x] = 0.5\\
&\mathbb{E}[\mathrm{hF1(Y}, \{1, 5\})|X=x] = 0.55
\end{align*}
 which shows that \roma{in a \textit{single path leaf label} setting it is strictly better to predict $\{1, 5\}$ when aiming at maximizing the hF1-score.}
With Proposition~\ref{prop:1} in mind, this simple example shows theoretically \textbf{the sub-optimality of the current state-of-the-art models inference methodology}.
As the optimal threshold is unknown, we need to design an evaluation framework which does not depend on an ad-hoc inference rule to avoid introducing non statistically grounded methods. Following recommendations given by~\citet{NEURIPS2022_727855c3}, we hence do away with inference rules and we construct precision-recall curves for hF1 by browsing all possible thresholds. From these curves, we compute the Area Under Curve (AUC).}

\section{Simple conditional loss-based \matt{methods}}

As a counterpart to the existing state-of-the-art consisting mainly of BCE-based approaches, we introduce several loss-based methods that incorporates local information, all relying on estimating \textbf{conditional probabilities}. 

\subsection{Conditional softmax cross-entropy}
\label{sec:cond_softmax}
As outlined in \hyperref[sec:prob_def]{Problem Definition}, we focus on methods that, given an input text $x$, produce an estimated distribution $\hat{\mathrm{P}}(\cdot|x)$ over $\mathcal{Y}$. We propose here to associate a modern text encoder to the conditional softmax~\cite{yolo9000}, 
which inherently incorporates the hierarchy structure by producing a hierarchy-coherent probability distribution, and coupling it with a cross-entropy loss. We detail in this section the modeling and training associated with it. Let us consider an input text $x$ with its corresponding label set $Y$; a text encoder is first used to produce an embedded representation $h_x \in \mathbb{R}^d$  of $x$. \\
\textbf{Conditional softmax}. The conditional softmax first maps $h_x$ to $s_x\in\mathbb{R}^{|\mathcal{Y}|}$ through a standard linear mapping: 
\begin{equation}
    s_x = Wh_x + b
    \label{eq:linear_map}
\end{equation}
where $W\in \mathbb{R}^{|\mathcal{Y}|\times d}$ and $b\in \mathbb{R}^{|\mathcal{Y}|}$.
Then, a softmax is applied to each brotherhood as follows: 
\begin{equation}
\label{eq:cond_softmax}
    \hat{\mathrm{P}}(y|x, \pi(y)) = \frac{\exp{s_x^{[y]}}}{\underset{z\in\mathcal{C}(\pi(y))}{\sum}\exp{s_x^{[z]}}}
\end{equation}
We recall that $\pi(y)$ denotes the parent node of node $y$, and $\mathcal{C}(\pi(y))$ represents the set of children of $\pi(y)$, which includes $y$. The term $s_x^{[y]}$ refers to the entry of $s_x$ associated with node $y$.  \\Hence, the logits $s_x$ are used to model the conditional probability of a node \textbf{given} its parent. For example, this could represent the probability of an instance $x$ to belong to the class \textit{Bulldog}, conditioned on it being a \textit{Dog}.

\noindent\textbf{Cross-entropy}. The contribution to the loss of the pair $(x, Y)$ is given by a standard leaf nodes cross-entropy (as if we were in a standard monolabel multiclass classification problem over leaf nodes). With our modelisation it can further be decomposed as: 
\begin{align}
\label{eq:loss_computation}
    l_{\mathrm{CSoft}}(x, Y) &= -\log\hat{\mathrm{P}}(y^{\text{leaf}}|x) \nonumber\\
    &=-\sum_{y \in Y}\log\hat{\mathrm{P}}(y|x, \pi(y))
\end{align}
where we denote $y^{\text{leaf}}$ the unique leaf node of $Y$.\\
\textbf{Outputted conditional distribution}. The probability of $y\in\mathcal{Y}$ is computed by a standard conditionality factorization :
\begin{equation*}
\label{eq:proba_decomp}
    \hat{\mathrm{P}}(y|x) = \prod_{z \in \mathcal{A}(y)}\hat{\mathrm{P}}(z|x, \pi(z)) 
\end{equation*}

\noindent\textbf{Motivations}. 
Contrary to BCE-based methods, this modelisation directly incorporates the hierarchy structure \romano{prior} of labels. 
Besides, the outputted probability distribution is coherent and exhaustive. 
It is more powerful than a leaf nodes softmax, as it decomposes the leaf probability estimation into several sub-problems. It is also computationally cheap, with a $\mathcal{O}(|\mathcal{Y}|)$ time complexity. 


\subsection{Logit-adjusted conditional softmax}
\matt{We then propose an enhanced version of the conditional softmax, in order to improve its robustness to data imbalance. This is particularly important for 
our newly introduced HWV dataset, which has around half of labels having less than $10$ instances in total. Our proposal is motivated by~\citet{zhou-etal-2020-hierarchy}, who suggest} that integrating the prior probability distribution in the model is relevant to the HTC task, which is confirmed by their experimental results. Their approach involves initializing (or fixing) the weights of the structure encoder using this pre-computed prior distribution. 
Hence, we draw inspiration from ~\citet{longtail} and introduce the logit-adjusted conditional softmax cross-entropy. 
Equation~\eqref{eq:cond_softmax} becomes: 
\begin{equation*}
    \hat{\mathrm{P}}(y|x, \pi(y)) = \frac{e^{s_x^{[y]} + \tau\log\nu(y|\pi(y))}}{\underset{z\in\mathcal{C}(\pi(y))}{\sum}e^{s_x^{[z]} + \tau\log\nu(z|\pi(z))}} 
\end{equation*}
where $\nu(y|\pi(y))$ is an estimation of  $\mathbb{P}(y|\pi(y))$\footnote{In practice, we estimate it by computing an empirical probability on train set for each label. It is not trainable.} and $\tau$ a hyperparameter. Equation~\eqref{eq:loss_computation} remains unchanged. 
Comprehensive details on the adaptation of the logit-adjusted softmax to our case, along with the theoretical justifications, are provided in Appendix~\ref{app:la}. We expect this loss to enhance performances on the under-represented classes.

\subsection{Conditional sigmoid binary cross-entropy}
\label{sec:cond_sigmoid}
In practice, several real-world datasets consistently used in recent literature to evaluate HTC models~\cite{rcv1, bgc} are multi-path. \matt{As the conditional softmax is not designed for multi-path labels, we propose to use a conditional sigmoid loss, introduced by \citet{brust2020integrating}.
It follows a similar intuition to the conditional softmax:} sigmoids are applied to each entry of $s_x$, modeling the conditional probability of the node given its parent. Hence, the contribution to the loss of a pair $(x, Y)$ is given by a \textbf{masked} cross-entropy\footnote{See Fig. 2b of \citet{brust2020integrating} for visual understanding of the mask}:
\begin{align*}
    l_{\mathrm{CSig}}(x, Y)&=-\underset{z \in Y}{\sum} \log(\hat{\mathrm{P}}(z|x, \pi(z))) \\+ &\underset{u \in \mathcal{C}(\pi(z))\backslash\lbrace{z}\rbrace}{\sum} \log(1- \hat{\mathrm{P}}(u|x, \pi(z)))
\end{align*}
\begin{proposition}
\label{prop:2}
    Let $x \in \mathcal{X}$, $Y\subset\mathcal{Y}$ and $W$ defined as per Equation~\ref{eq:linear_map} then 
\begin{equation*}
    \frac{\partial l_{\mathrm{CSoft}}(x, Y)}{\partial W} = \frac{\partial l_{\mathrm{CSig}}(x, Y)}{\partial W}
\end{equation*}
\end{proposition}
Proof can be found in Appendix~\ref{sec:soft_sig}.
While the conditional sigmoid was not motivated by theoretical arguments in~\citet{brust2020integrating}, Proposition~\ref{prop:2} proves that gradients computed for this loss and the conditional softmax cross-entropy loss are equivalent.
This loss then allows to deal with both multi-path and non-exhaustive datasets while having similar properties to conditional softmax.\footnote{However, no logit-adjusted version of it can be properly derived.} 

\section{Experimental settings}

In this section, we introduce \matt{the existing} datasets and models \matt{we experiment with; we also present our new dataset, Hierarchical WikiVitals (HWV).} 

\subsection{Datasets}
\label{sec:datasets}
We will verify the performance of our proposed approaches versus baselines and recent state-of-the-art models on hierarchical metrics on three widely used datasets in the HTC literature, which is mainly applied to English data: Web-of-Science (WOS)~\cite{WOS}, RCV1-V2~\citep{rcv1} and BGC~\cite{bgc}. \matt{Data statistics are displayed in Table~\ref{tab:stats_data}: those datasets have in common a relatively large number of training samples, a sizable number of nodes, and a low depth of the label structure.} \matt{We contribute to HTC benchmarking by releasing Hierarchical WikiVitals, which we aim to present a more difficult challenge.} 

\begin{table}[!t]
    \centering
    \resizebox{\columnwidth}{!}{
    \begin{tabular}
    {lcccc}
        Dataset & \makecell{Train/Val/Test} & \makecell{\#nodes \\ (\#leaves)}  & \makecell{\#nodes \\per level} & \makecell{Avg. \#labels\\ per sample} 
        \\ 
        \hline
        HWV (SPL) & \makecell{6,408/1,602 \\ 2,003} & \makecell{1186 \\ (953)} & \makecell{11-109-381-\\437-244-4} & 3.7 
        \\
        \hline
        WOS (SPL) & \makecell{30,070/7,518 \\ 9,397}  & \makecell{141 \\(134)} & 7-134 & 2.0 
        \\
        \hline
        RCV1 (MP) & \makecell{23,149/ - \\ 781,265} & \makecell{103\\ (82)} & 4-55-43-1 & 3.2
        \\
        \hline
        BGC (MP) & \makecell{58,715/14,785 \\ 18,394} & \makecell{146 \\ (120)} & 7-46-77-16 & 3.0 
        \\
    \end{tabular}
    }
    \caption{Key statistics of the selected datasets. \textbf{SPL} indicates that the dataset enters the \textit{single path leaf labels} setting, and \textbf{MP} that it is multi-path; $d$ represents the maximum depth of the label hierarchy.}
    \label{tab:stats_data}
\end{table}

\paragraph{HWV Dataset} \roma{Texts are extracted from the abstracts of the \textit{vital} articles of Wikipedia, level 4 \footnote{\url{https://en.wikipedia.org/wiki/Wikipedia:Vital_articles/Level/4}} as of June 2021. This project involves a handmade hierarchical categorization of the selected articles, which are themselves put through high scrutiny with respect to their quality. The resulting dataset is a \textit{single path leaf label} dataset, \matt{a constraint only fulfilled by WOS.}} 
As the number of nodes and the depth of the hierarchy are higher than for the previously cited datasets, HWV is much more challenging. \rom{It is also characterized by a very imbalanced label distribution with $\sim 50\%$ of labels having less than $10$ examples in the whole dataset. We show in Figure~\ref{fig:enter-label} three observations from our new dataset, illustrating how much leaf nodes depth can vary (ranging from 2 to 6)}. 
Comprehensive details regarding the building process of the quality of data of HWV are provided in Appendix~\ref{sec:hwv}.

\subsection{Models}
We propose to compare very different HTC models, ranging from simple baselines to the most recent state-of-the-art approaches. For fair comparison between them, we use a pre-trained BERT\footnote{\url{https://huggingface.co/bert-base-uncased}} model~\cite{devlin2019bert} as text encoder, adopting the standard [CLS] representation as $h_x$ for every model. 
We list below all the different models evaluated. \textbf{BERT + BCE} is the simplest baseline, treating the problem as a multi-label task, without using any information from the hierarchical structure of labels. \textbf{BERT + Leaf Softmax} outputs a distribution over leaves, and hence is only fitted for {single-path leaf label} settings. \textbf{BERT + CHAMP} implements the penalization of false positives based on their shortest-path distance to the ground label set in the tree~\citep{vaswani2022champ}. \textbf{BERT + Conditional \{Softmax, logit-adjusted Softmax, Sigmoid\}} are \matt{our proposed methods, }
detailed in Section~\ref{sec:cond_softmax}. \textbf{Hitin}~\citep{zhu2023hitin}, \textbf{HBGL}~\citep{jiang-etal-2022-exploiting}, \textbf{HGCLR}~\citep{wang2022incorporating} are among the most recent models, proposing respectively to separately encode the label hierarchy in an efficient manner, to incorporate both global and local information when encoding the label hierarchy, by considering subgraphs, and to use contrastive learning and exploiting the label hierarchy to create plausible corrupted examples. 

\subsection{Training details}

We use \texttt{bert-base-uncased} model from the transformers library~\cite{wolf-etal-2020-transformers} as text encoder (110M parameters). Our implementation is based on Hitin.\footnote{\url{https://github.com/Rooooyy/HiTIN}} Each of our baselines is trained for 20 epochs on a V100 GPU of 32GB with a batch size of 16. We used an AdamW optimizer with initial learning rate of $2\cdot10^{-5}$ and with a warmup period of $10\%$ of the training steps. For HBGL\footnote{\url{https://github.com/kongds/HBGL}}, Hitin and HGCLR\footnote{\url{https://github.com/wzh9969/contrastive-htc}}, we rely on implementation guidelines to conduct experiments. For datasets not used in the original papers, we performed a grid-search hyperparameter optimization. 
Our results are derived from averaging over four separate training runs, each initialized with distinct random seeds, ensuring the robustness and fairness of our evaluation methodology.

\begin{table*}[!h]        
    \centering
    \resizebox{\textwidth}{!}{
    \begin{tabular}{|l|c|c|c|c||c|c|c|c|}
    \hline
        & \multicolumn{4}{c||}{HWV} & \multicolumn{4}{c|}{WOS}\\ 
        \hline
        \multirow{2}{*}{Method} & Hamming L.  & \multicolumn{2}{c|}{F1-score (in \%) $\uparrow$} & \multirow{2}{*}{\makecell{hF1 AUC \\(in \%) $\uparrow$}} & Hamming L. & \multicolumn{2}{c|}{F1-score (in \%) $\uparrow$} & \multirow{2}{*}{\makecell{hF1 AUC \\(in \%) $\uparrow$}} \\ \cline{3-4} \cline{7-8}
         & (in \textperthousand) $\downarrow$ & micro & macro &   & (in \textperthousand) $\downarrow$ & micro & macro & \\ 
        \hline
        BCE   & 0.854$_{\pm 0.010}$ & 85.86$_{\pm 0.15}$ & 45.56$_{\pm 0.58}$& 89.23$_{\pm 0.13}$  & 3.627$_{\pm 0.015}$& 87.03$_{\pm 0.05}$ & 81.19$_{\pm 0.12}$ & \textbf{89.18$_{\pm 0.10}$} \\
        CHAMP  & 0.786$_{\pm 0.009}$ & 87.14$_{\pm 0.15}$& 50.90$_{\pm 0.24}$& 89.87$_{\pm 0.19}$ & 3.637$_{\pm 0.037}$& 87.01$_{\pm 0.13}$ & 81.23$_{\pm 0.18}$ & 88.74$_{\pm 0.08}$ \\
        \hline
        HBGL  & - & - & - & - & \textbf{3.584$_{\pm 0.027}$} & \textbf{87.22$_{\pm 0.10}$} & \textbf{81.86$_{\pm 0.19}$} &89.00$_{\pm 0.10}$\\
        HGCLR  & 0.922$_{\pm 0.020}$ & 84.92$_{\pm 0.37}$& 44.89$_{\pm 1.38}$ & 88.35$_{\pm 0.35}$  & 3.727$_{\pm 0.077}$ & 86.63$_{\pm 0.27}$ & 80.04$_{\pm 0.45}$ & \textbf{89.23$_{\pm 0.22}$} \\
        HITIN   & \textbf{0.776$_{\pm 0.006}$} & \textbf{87.49$_{\pm 0.08}$}& 51.73$_{\pm 0.42}$ & 90.72$_{\pm 0.16}$ & 3.655$_{\pm 0.028}$ & 87.05$_{\pm 0.10}$ & 81.49$_{\pm 0.06}$ & 88.92$_{\pm 0.04}$\\
        \hline
        Leaf Softmax  & 0.950$_{\pm 0.036}$ & 84.79$_{\pm 0.57}$ & 51.49$_{\pm 0.52}$& 88.55$_{\pm 0.47}$ & 3.987$_{\pm 0.059}$ & 85.91$_{\pm 0.25}$ & 80.02$_{\pm 0.29}$ & 88.62$_{\pm 0.08}$\\
        Conditional Sigmoid & 0.801$_{\pm 0.011}$ & 87.01$_{\pm 0.19}$ & 52.27$_{\pm 0.82}$ & 90.40$_{\pm 0.17}$ & 3.692$_{\pm 0.067}$ & 86.86$_{\pm 0.23}$ & 81.07$_{\pm 0.30}$ & 88.78$_{\pm 0.17}$ \\
        Conditional Softmax  & 0.788$_{\pm 0.015}$ & \textbf{87.49$_{\pm 0.10}$}& 53.79$_{\pm 0.65}$& \textbf{90.94$_{\pm 0.09}$} & 3.869$_{\pm 0.086}$ & 86.27$_{\pm 0.17}$ & 80.25$_{\pm 0.33}$ & 88.77$_{\pm 0.07}$ \\
        Cond. Softmax + LA  (ours) & \textbf{0.782$_{\pm 0.004}$} & \textbf{87.51$_{\pm 0.07}$} & \textbf{54.39$_{\pm 0.58}$} & \textbf{90.97$_{\pm 0.05}$} & 3.837$_{\pm 0.038}$& 86.35$_{\pm 0.12}$ & 80.11$_{\pm 0.26}$ & 88.90$_{\pm 0.10}$\\
        \hline
    \end{tabular}
    }
    \caption{Performance evaluation metrics (and $95\%$ confidence interval) on the test sets of the WOS and HWV datasets for the implemented models. 
    Best results for each metric are highlighted in bold. The HBGL model was too large to fit in the memory of a 32GB GPU on the HWV dataset.}
    \label{tab:res_single_path}
\end{table*}

\begin{table*}[!h]
    \centering
    \resizebox{\textwidth}{!}{
    \begin{tabular}{|l|c|c|c|c||c|c|c|c|}
    \hline
        & \multicolumn{4}{c||}{RCV1} & \multicolumn{4}{c|}{BGC}\\ 
        \hline
        \multirow{2}{*}{Method} & Hamming L.  & \multicolumn{2}{c|}{F1-score (in \%) $\uparrow$} & \multirow{2}{*}{\makecell{hF1 AUC \\(in \%) $\uparrow$}} & Hamming L. & \multicolumn{2}{c|}{F1-score (in \%) $\uparrow$} & \multirow{2}{*}{\makecell{hF1 AUC \\(in \%) $\uparrow$}} \\ \cline{3-4} \cline{7-8}
         & (in \textperthousand) $\downarrow$ & micro & macro &   & (in \textperthousand) $\downarrow$ & micro & macro & \\ 
        \hline
        BCE  & 8.225$_{\pm 0.148}$ & 86.65$_{\pm 0.30}$ & 66.47$_{\pm 1.49}$ & \textbf{93.66$_{\pm 0.19}$} & \textbf{7.788$_{\pm 0.071}$} & \textbf{80.51$_{\pm 0.21}$} & 62.33$_{\pm 1.36}$& \textbf{90.26$_{\pm 0.29}$}\\
        CHAMP  & 8.565$_{\pm 0.234}$& 85.93$_{\pm 0.66}$ & 62.86$_{\pm 3.64}$ & 93.12$_{\pm 0.33}$ & \textbf{7.775$_{\pm 0.081}$} & \textbf{80.54$_{\pm 0.20}$} & 63.58$_{\pm 0.49}$ & \textbf{90.19$_{\pm 0.22}$}\\
        \hline
        HBGL  & \textbf{8.122$_{\pm 0.071}$} & \textbf{87.11$_{\pm 0.12}$} & \textbf{70.20$_{\pm 0.33}$} & 93.35$_{\pm 0.14}$ & 8.092$_{\pm 0.045}$ & 80.19$_{\pm 0.11}$ & \textbf{65.94$_{\pm 0.18}$} & 88.08$_{\pm 0.10}$ \\
        HGCLR  & 8.761$_{\pm 0.276}$ & 86.11$_{\pm 0.26}$ & 67.49$_{\pm 0.61}$ & 93.27$_{\pm 0.14}$& 8.054$_{\pm 0.171}$ & 80.16$_{\pm 0.29}$ & 63.58$_{\pm 0.40}$ & 89.81$_{\pm 0.17}$\\
        HITIN  & 8.583$_{\pm 0.188}$ & 85.72$_{\pm 0.60}$ & 60.00$_{\pm 5.15}$ & 93.04$_{\pm 0.24}$  & 7.981$_{\pm 0.096}$ & \textbf{80.36$_{\pm 0.21}$} & 61.62$_{\pm 1.47}$ &\textbf{ 90.08$_{\pm 0.16}$}\\
        \hline
        Conditional Sigmoid  & 8.652$_{\pm 0.316}$ & 85.77$_{\pm 0.71}$ & 63.90$_{\pm 2.45}$ & 93.23$_{\pm 0.36}$ & 7.954$_{\pm 0.202}$ & 80.24$_{\pm 0.46}$& 62.65$_{\pm 0.64}$& \textbf{90.07$_{\pm 0.40}$}\\
        \hline
    \end{tabular}
    }
    \caption{Performance evaluation metrics (and $95\%$ confidence interval) on the test sets of the RCV1 and BGC datasets for the implemented models. 
    Best results for each metric are highlighted in bold.}
    \label{tab:res_multi_path}
\end{table*}
\section{Results and Analysis}

\begin{figure}[!t]
    \centering
    \includegraphics[width=1\linewidth]{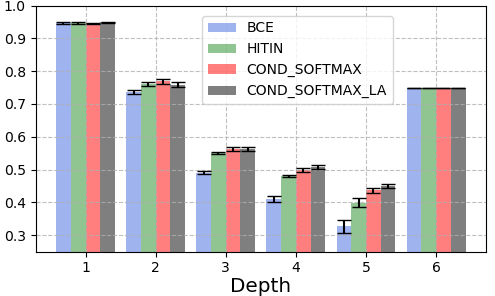} 
    \caption{Averaged Macro F1-Scores on the test set per depth for different models and for the HWV dataset. The error bars represent a $95\%$ confidence interval.}
    \label{fig:macro_fa_depth}
\end{figure}

\begin{figure}[!t]
    \centering
    \includegraphics[width=\linewidth]{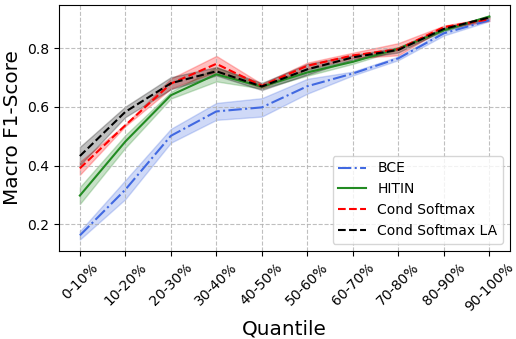} 
    \caption{Averaged Macro F1-Scores on the test set by quantiles of label counts distribution in the training set for different models and for the HWV dataset. The shaded regions represent a $95\%$ confidence interval.}
    \label{fig:macro_fa_quantile_left}
\end{figure}

We start our investigation by evaluating models on our newly proposed dataset, HWV. \matt{Results are shown in Table~\ref{tab:res_single_path}}. Unfortunately, the HBGL architecture could not run for HWV, requiring memory above the capacity of our GPUs. 
\textbf{\matt{On this dataset, we note} the overall superiority of our newly introduced logit-adjusted conditional softmax loss and its vanilla version.} 
The latest models fail to obtain the best results, which is surprising given the complex hierarchy and label imbalance. We hence emit the hypothesis that while \textit{hierarchy-aware} models were proven useful on simpler datasets, they fail to capture that complexity on HWV. To investigate why it performs better, we display in Figures~\ref{fig:macro_fa_depth} \& \ref{fig:macro_fa_quantile_left}  averaged macro F1-scores over classes. Figure~\ref{fig:macro_fa_depth} corresponds to averages of scores based on label depth: \romano{we observe 
that the higher the depth the higher the improvement brought by conditional softmax and its logit-adjusted version is (except for depth 6 which has only 4 classes inside). Figure~\ref{fig:macro_fa_quantile_left} seems to hint that the improvement of the logit-adjusted conditional softmax vs. a vanilla conditional softmax lies in its ability to correctly classify \textit{under-represented} classes. Until the third decile of the label count distribution, our newly introduced method is statistically better. We could have expected such a result, as this loss was specifically designd to deal with label imbalance (see Appendix \ref{app:la}). Obviously, depth is strongly correlated with \textit{under-representation} of labels. We then conduct an ablation study with respect to the label hierarchy, by cutting the HWV hierarchy at depth 2. By doing so, the hierarchy becomes shallow and the label \textit{imbalance} remains. Table \ref{tab:hwv_cutted} presents the results obtained from this modified dataset. In this scenario, state-of-the-art models catch up with our conditional softmax losses and {Hitin} reclaim a marginal lead across all metrics. Furthermore, we observe that our logit-adjusted conditional softmax remains better than the vanilla conditional softmax, especially on \textit{macro} F1-score. These two observations allow us to refine our conclusions. First, the superiority of the vanilla conditional softmax on HWV vs. recent state-of-the-art methods seems to stem from the {hierarchy complexity}: \textbf{a conditional modelisation allows to better classify deep classes}. Second, the logit-adjusted version proves to be useful {in presence of label imbalance} as we can see with \textit{macro} F1-score metrics, which are statistically better than the vanilla version in both versions of HWV dataset.}\\
On WOS, simpler baselines reach remarkable results. Despite the marginal superiority of HBGL, it is noteworthy that the
\mat{\textbf{BERT+BCE model is in the top performances across all metrics}, while not using label hierarchy information.}
On this dataset, our new method, while competitive, lags behind. 
\romano{These results are coherent with conclusions drawn with HWV dataset} : the WOS dataset has a low complexity, both in terms of depth (maximum depth of 2) and distribution of labels (only one class has less than 40 examples in the dataset). \\
On multi-path datasets, our observations align closely with what we noticed on WOS: we observe in Table~\ref{tab:res_multi_path} that a straightforward BCE loss consistently yields great results across datasets and metrics. \matt{Hierarchical metrics clearly highlight this phenomenon. In fact, model rankings in multi-label F1 scores and hierarchical F1 scores only keep consistent for HWV: for the three other datasets, the \textbf{structure-aware threshold-independent metrics put the BCE baseline to the top}.}

\begin{table}[!t]
    \centering
    \resizebox{0.48\textwidth}{!}{
   \begin{tabular}{|l|c|c|c|c|c|}
    \hline
        & \multicolumn{4}{c|}{HWV (depth 2)} \\
        \hline
        \multirow{2}{*}{Method} & Hamming L.  & \multicolumn{2}{c|}{F1-score (in \%) $\uparrow$} & \multirow{2}{*}{\makecell{hF1 AUC \\(in \%) $\uparrow$}} \\ \cline{3-4}
        & (in \textperthousand) $\downarrow$ & micro & macro &   \\ 
        \hline
        BCE  & 2.367$_{\pm 0.030}$  & \textbf{92.89$_{\pm 0.06}$} & 78.42$_{\pm 0.31}$& \textbf{94.67$_{\pm 0.15}$}  \\
        HITIN  & \textbf{2.316$_{\pm 0.068}$} & \textbf{93.05$_{\pm 0.20}$} & \textbf{79.59$_{\pm 0.43}$} & \textbf{94.79$_{\pm 0.18}$}  \\
        \hline
        Cond. Soft.  & 2.450$_{\pm 0.087}$ & 92.65$_{\pm 0.26}$ & 78.40$_{\pm 1.06}$ & \textbf{94.73$_{\pm 0.18}$}\\
        Cond. S. L.A.  & 2.432$_{\pm 0.072}$ & \textbf{92.89$_{\pm 0.22}$} & \textbf{79.38$_{\pm 0.26}$} & \textbf{94.77$_{\pm 0.18}$} \\
        \hline
    \end{tabular}
    }
    \caption{Performance evaluation metrics (and $95\%$ confidence interval) on the test sets of the cutted HWV dataset for the implemented models. 
    Best results for each metric are highlighted in bold.}
    \label{tab:hwv_cutted}
\end{table}
We believe those results allow us to draw two main lessons: first, that \matt{hierarchical metrics bring useful insights on HTC evaluation, and are necessary to properly evaluate models on their capacity to encode label structure, which our results show to be lacking.} 
\matt{Second, that} when used on a more challenging dataset, state-of-the-art {hierarchy-aware} HTC models are less able to integrate that complex hierarchical information into their prediction than a simple model trained with conditional softmax~cross-entropy. 


\section{Conclusion}
In this paper, we come back upon recent progress in HTC, and propose to investigate its evaluation. To do so, we begin by showing the 
limitations of the inference and metrics that are commonly used in the recent literature. We instead propose to use existing hierarchical metrics, and an associated inference method. 
Then, we introduce a new and challenging dataset, Hierarchical WikiVitals; our experiments show that recent sophisticated {hierarchy-aware} models have trouble integrating hierarchy information \matt{in any better way than simple baselines. We finally propose simple hierarchical losses,} 
able to better integrate hierarchy information on our dataset. In the future, we plan to investigate the inference mechanism for hierarchical metrics, through which we will aim to make a direct contribution~to improving models on HTC tasks. 



\bibliography{acl_latex}

\begin{thebibliography}{51}
\expandafter\ifx\csname natexlab\endcsname\relax\def\natexlab#1{#1}\fi

\bibitem[{Aly et~al.(2019)Aly, Remus, and Biemann}]{bgc}
Rami Aly, Steffen Remus, and Chris Biemann. 2019.
\newblock \href {https://doi.org/10.18653/v1/P19-2045} {Hierarchical multi-label classification of text with capsule networks}.
\newblock In \emph{Proceedings of the 57th Annual Meeting of the Association for Computational Linguistics: Student Research Workshop}, pages 323--330, Florence, Italy. Association for Computational Linguistics.

\bibitem[{Amigo and Delgado(2022)}]{amigo-delgado-2022-evaluating}
Enrique Amigo and Agust{\'\i}n Delgado. 2022.
\newblock \href {https://doi.org/10.18653/v1/2022.acl-long.399} {Evaluating extreme hierarchical multi-label classification}.
\newblock In \emph{Proceedings of the 60th Annual Meeting of the Association for Computational Linguistics (Volume 1: Long Papers)}, pages 5809--5819, Dublin, Ireland. Association for Computational Linguistics.

\bibitem[{Banerjee et~al.(2019)Banerjee, Akkaya, Perez-Sorrosal, and Tsioutsiouliklis}]{banerjee-etal-2019-hierarchical}
Siddhartha Banerjee, Cem Akkaya, Francisco Perez-Sorrosal, and Kostas Tsioutsiouliklis. 2019.
\newblock \href {https://doi.org/10.18653/v1/P19-1633} {Hierarchical transfer learning for multi-label text classification}.
\newblock In \emph{Proceedings of the 57th Annual Meeting of the Association for Computational Linguistics}, pages 6295--6300, Florence, Italy. Association for Computational Linguistics.

\bibitem[{Berger(1985)}]{berger1985statistical}
James~O. Berger. 1985.
\newblock \href {https://doi.org/10.1007/978-1-4757-4286-2} {\emph{Statistical Decision Theory and Bayesian Analysis}}, 2nd edition.
\newblock Springer Series in Statistics. Springer, New York.

\bibitem[{Bertinetto et~al.(2020)Bertinetto, Mueller, Tertikas, Samangooei, and Lord}]{bertinetto2020}
Luca Bertinetto, Romain Mueller, Konstantinos Tertikas, Sina Samangooei, and Nicholas~A. Lord. 2020.
\newblock Making better mistakes: Leveraging class hierarchies with deep networks.
\newblock In \emph{IEEE/CVF Conference on Computer Vision and Pattern Recognition (CVPR)}.

\bibitem[{Bi and Kwok(2015)}]{7118216}
Wei Bi and Jame~T. Kwok. 2015.
\newblock \href {https://doi.org/10.1109/TKDE.2015.2441707} {Bayes-optimal hierarchical multilabel classification}.
\newblock \emph{IEEE Transactions on Knowledge and Data Engineering}, 27(11):2907--2918.

\bibitem[{Brust and Denzler(2020)}]{brust2020integrating}
Clemens-Alexander Brust and Joachim Denzler. 2020.
\newblock Integrating domain knowledge: Using hierarchies to improve deep classifiers.
\newblock In \emph{Pattern Recognition}, pages 3--16, Cham. Springer International Publishing.

\bibitem[{Chen et~al.(2020)Chen, Huang, Xiao, Cai, and Jing}]{chen2019hyperbolic}
Boli Chen, Xin Huang, Lin Xiao, Zixin Cai, and Liping Jing. 2020.
\newblock Hyperbolic interaction model for hierarchical multi-label classification.
\newblock In \emph{Proceedings of the AAAI conference on artificial intelligence}, volume~34, pages 7496--7503.

\bibitem[{Chen et~al.(2021)Chen, Ma, Lin, and Yan}]{HiMatch}
Haibin Chen, Qianli Ma, Zhenxi Lin, and Jiangyue Yan. 2021.
\newblock \href {https://doi.org/10.18653/v1/2021.acl-long.337} {Hierarchy-aware label semantics matching network for hierarchical text classification}.
\newblock In \emph{Proceedings of the 59th Annual Meeting of the Association for Computational Linguistics and the 11th International Joint Conference on Natural Language Processing (Volume 1: Long Papers)}, pages 4370--4379, Online. Association for Computational Linguistics.

\bibitem[{Collell et~al.(2017)Collell, Prelec, and Patil}]{proof2}
Guillem Collell, Drazen Prelec, and Kaustubh Patil. 2017.
\newblock \href {http://arxiv.org/abs/1606.08698} {Reviving threshold-moving: a simple plug-in bagging ensemble for binary and multiclass imbalanced data}.
\newblock ArXiv preprint.

\bibitem[{Costa et~al.(2007)Costa, Lorena, Carvalho, and Freitas}]{review_metrics}
Eduardo Costa, Ana Lorena, Andre Carvalho, and Alex Freitas. 2007.
\newblock A review of performance evaluation measures for hierarchical classifiers.
\newblock \emph{AAAI Workshop - Technical Report}.

\bibitem[{Dembczy{\'n}ski et~al.(2012)Dembczy{\'n}ski, Waegeman, Cheng, and H{\"u}llermeier}]{hammingloss}
Krzysztof Dembczy{\'n}ski, Willem Waegeman, Weiwei Cheng, and Eyke H{\"u}llermeier. 2012.
\newblock On label dependence and loss minimization in multi-label classification.
\newblock \emph{Machine Learning}, 88:5--45.

\bibitem[{Deng et~al.(2021)Deng, Peng, He, Li, and Yu}]{deng2021htcinfomax}
Zhongfen Deng, Hao Peng, Dongxiao He, Jianxin Li, and Philip Yu. 2021.
\newblock \href {https://doi.org/10.18653/v1/2021.naacl-main.260} {{HTCI}nfo{M}ax: A global model for hierarchical text classification via information maximization}.
\newblock In \emph{Proceedings of the 2021 Conference of the North American Chapter of the Association for Computational Linguistics: Human Language Technologies}, pages 3259--3265, Online. Association for Computational Linguistics.

\bibitem[{Devlin et~al.(2019)Devlin, Chang, Lee, and Toutanova}]{devlin2019bert}
Jacob Devlin, Ming-Wei Chang, Kenton Lee, and Kristina Toutanova. 2019.
\newblock \href {https://doi.org/10.18653/v1/N19-1423} {{BERT}: Pre-training of deep bidirectional transformers for language understanding}.
\newblock In \emph{Proceedings of the 2019 Conference of the North {A}merican Chapter of the Association for Computational Linguistics: Human Language Technologies, Volume 1 (Long and Short Papers)}, pages 4171--4186, Minneapolis, Minnesota. Association for Computational Linguistics.

\bibitem[{Gopal and Yang(2013)}]{recursive_reguluation_Gopal_2013}
Siddharth Gopal and Yiming Yang. 2013.
\newblock \href {https://doi.org/10.1145/2487575.2487644} {Recursive regularization for large-scale classification with hierarchical and graphical dependencies}.
\newblock In \emph{Proceedings of the 19th ACM SIGKDD International Conference on Knowledge Discovery and Data Mining}, KDD '13, page 257–265, New York, NY, USA. Association for Computing Machinery.

\bibitem[{Huang et~al.(2019)Huang, Chen, Liu, Chen, Huang, Liu, Zhao, Zhang, and Wang}]{huang}
Wei Huang, Enhong Chen, Qi~Liu, Yuying Chen, Zai Huang, Yang Liu, Zhou Zhao, Dan Zhang, and Shijin Wang. 2019.
\newblock Hierarchical multi-label text classification: An attention-based recurrent network approach.
\newblock In \emph{Proceedings of the 28th ACM international conference on information and knowledge management}, pages 1051--1060.

\bibitem[{Ji et~al.(2023)Ji, Lian, Gao, and Wang}]{C_F1}
Ke~Ji, Yixin Lian, Jingsheng Gao, and Baoyuan Wang. 2023.
\newblock \href {https://doi.org/10.18653/v1/2023.acl-long.164} {Hierarchical verbalizer for few-shot hierarchical text classification}.
\newblock In \emph{Proceedings of the 61st Annual Meeting of the Association for Computational Linguistics (Volume 1: Long Papers)}, pages 2918--2933, Toronto, Canada. Association for Computational Linguistics.

\bibitem[{Jiang et~al.(2022)Jiang, Wang, Sun, Chen, Zhuang, and Yang}]{jiang-etal-2022-exploiting}
Ting Jiang, Deqing Wang, Leilei Sun, Zhongzhi Chen, Fuzhen Zhuang, and Qinghong Yang. 2022.
\newblock \href {https://doi.org/10.18653/v1/2022.emnlp-main.268} {Exploiting global and local hierarchies for hierarchical text classification}.
\newblock In \emph{Proceedings of the 2022 Conference on Empirical Methods in Natural Language Processing}, pages 4030--4039, Abu Dhabi, United Arab Emirates. Association for Computational Linguistics.

\bibitem[{Kiritchenko et~al.(2006)Kiritchenko, Matwin, Nock, and Famili}]{hierf1score}
Svetlana Kiritchenko, Stan Matwin, Richard Nock, and A.~Fazel Famili. 2006.
\newblock Learning and evaluation in the presence of class hierarchies: Application to text categorization.
\newblock In \emph{Advances in Artificial Intelligence}, pages 395--406, Berlin, Heidelberg. Springer Berlin Heidelberg.

\bibitem[{Kosmopoulos et~al.(2014)Kosmopoulos, Partalas, Gaussier, Paliouras, and Androutsopoulos}]{evaluation_metrics}
Aris Kosmopoulos, Ioannis Partalas, Eric Gaussier, Georgios Paliouras, and Ion Androutsopoulos. 2014.
\newblock \href {https://doi.org/10.1007/s10618-014-0382-x} {Evaluation measures for hierarchical classification: a unified view and novel approaches}.
\newblock \emph{Data Mining and Knowledge Discovery}, 29(3):820--865.

\bibitem[{Kowsari et~al.(2018)Kowsari, Brown, Heidarysafa, Jafari~Meimandi, Gerber, and Barnes}]{WOS}
Kamran Kowsari, Donald Brown, Mojtaba Heidarysafa, Kiana Jafari~Meimandi, Matthew Gerber, and Laura Barnes. 2018.
\newblock \href {https://doi.org/10.17632/9rw3vkcfy4.6} {Web of science dataset}.

\bibitem[{Lewis et~al.(2004)Lewis, Yang, Rose, and Li}]{rcv1}
D.~D. Lewis, Y.~Yang, T.~G. Rose, and F.~Li. 2004.
\newblock \href {http://www.jmlr.org/papers/volume5/lewis04a/lewis04a.pdf} {Rcv1: A new benchmark collection for text categorization research}.
\newblock \emph{Journal of Machine Learning Research}, 5(Apr):361--397.

\bibitem[{Lyubinets et~al.(2018)Lyubinets, Boiko, and Nicholas}]{bugs}
Volodymyr Lyubinets, Taras Boiko, and Deon Nicholas. 2018.
\newblock \href {https://doi.org/10.1109/DSMP.2018.8478511} {Automated labeling of bugs and tickets using attention-based mechanisms in recurrent neural networks}.
\newblock In \emph{2018 IEEE Second International Conference on Data Stream Mining and Processing (DSMP)}, pages 271--275.

\bibitem[{Mao et~al.(2019)Mao, Tian, Han, and Ren}]{mao-etal-2019-hierarchical}
Yuning Mao, Jingjing Tian, Jiawei Han, and Xiang Ren. 2019.
\newblock \href {https://doi.org/10.18653/v1/D19-1042} {Hierarchical text classification with reinforced label assignment}.
\newblock In \emph{Proceedings of the 2019 Conference on Empirical Methods in Natural Language Processing and the 9th International Joint Conference on Natural Language Processing (EMNLP-IJCNLP)}, pages 445--455, Hong Kong, China. Association for Computational Linguistics.

\bibitem[{Menon et~al.(2013)Menon, Narasimhan, Agarwal, and Chawla}]{proof1}
Aditya Menon, Harikrishna Narasimhan, Shivani Agarwal, and Sanjay Chawla. 2013.
\newblock \href {https://proceedings.mlr.press/v28/menon13a.html} {On the statistical consistency of algorithms for binary classification under class imbalance}.
\newblock In \emph{Proceedings of the 30th International Conference on Machine Learning}, volume~28 of \emph{Proceedings of Machine Learning Research}, pages 603--611, Atlanta, Georgia, USA. PMLR.

\bibitem[{Menon et~al.(2021)Menon, Veit, Rawat, Jain, Jayasumana, and Kumar}]{longtail}
Aditya~Krishna Menon, Andreas Veit, Ankit~Singh Rawat, Himanshu Jain, Sadeep Jayasumana, and Sanjiv Kumar. 2021.
\newblock Long-tail learning via logit adjustment.
\newblock In \emph{International Conference on Learning Representations (ICLR) 2021}.

\bibitem[{Redmon and Farhadi(2017)}]{yolo9000}
Joseph Redmon and Ali Farhadi. 2017.
\newblock Yolo9000: Better, faster, stronger.
\newblock In \emph{Proceedings of the IEEE Conference on Computer Vision and Pattern Recognition (CVPR)}.

\bibitem[{Sandhaus(2008)}]{nyt}
Evan Sandhaus. 2008.
\newblock The new york times annotated corpus.
\newblock \emph{Linguistic Data Consortium}, 6(12):e26752.

\bibitem[{Shimura et~al.(2018)Shimura, Li, and Fukumoto}]{shimura-etal-2018-hft}
Kazuya Shimura, Jiyi Li, and Fumiyo Fukumoto. 2018.
\newblock \href {https://doi.org/10.18653/v1/D18-1093} {{HFT}-{CNN}: Learning hierarchical category structure for multi-label short text categorization}.
\newblock In \emph{Proceedings of the 2018 Conference on Empirical Methods in Natural Language Processing}, pages 811--816, Brussels, Belgium. Association for Computational Linguistics.

\bibitem[{Silla and Freitas(2011)}]{silla_freitas}
Carlos Silla and Alex Freitas. 2011.
\newblock \href {https://doi.org/10.1007/s10618-010-0175-9} {A survey of hierarchical classification across different application domains}.
\newblock \emph{Data Mining and Knowledge Discovery}, 22:31--72.

\bibitem[{Song et~al.(2012)Song, Son, Noh, Park, and Lee}]{song-etal-2012-cost}
Hyun-Je Song, Jeong-Woo Son, Tae-Gil Noh, Seong-Bae Park, and Sang-Jo Lee. 2012.
\newblock \href {https://aclanthology.org/P12-1108} {A cost sensitive part-of-speech tagging: Differentiating serious errors from minor errors}.
\newblock In \emph{Proceedings of the 50th Annual Meeting of the Association for Computational Linguistics (Volume 1: Long Papers)}, pages 1025--1034, Jeju Island, Korea. Association for Computational Linguistics.

\bibitem[{Song et~al.(2023)Song, Wang, and Yang}]{song-etal-2023-peer}
Junru Song, Feifei Wang, and Yang Yang. 2023.
\newblock \href {https://doi.org/10.18653/v1/2023.acl-long.207} {Peer-label assisted hierarchical text classification}.
\newblock In \emph{Proceedings of the 61st Annual Meeting of the Association for Computational Linguistics (Volume 1: Long Papers)}, pages 3747--3758, Toronto, Canada. Association for Computational Linguistics.

\bibitem[{Sun and Lim(2001)}]{sunlin2001}
Aixin Sun and Ee-Peng Lim. 2001.
\newblock \href {https://doi.org/10.1109/ICDM.2001.989560} {Hierarchial text classification and evaluation}.
\newblock pages 521--528.

\bibitem[{Valmadre(2022)}]{NEURIPS2022_727855c3}
Jack Valmadre. 2022.
\newblock \href {https://proceedings.neurips.cc/paper_files/paper/2022/file/727855c31df8821fd18d41c23daebf10-Paper-Conference.pdf} {Hierarchical classification at multiple operating points}.
\newblock In \emph{Advances in Neural Information Processing Systems}, volume~35, pages 18034--18045. Curran Associates, Inc.

\bibitem[{Vaswani et~al.(2022)Vaswani, Aggarwal, Netrapalli, and Hegde}]{vaswani2022champ}
Ashwin Vaswani, Gaurav Aggarwal, Praneeth Netrapalli, and Narayan~G Hegde. 2022.
\newblock \href {http://arxiv.org/abs/2206.08653} {All mistakes are not equal: Comprehensive hierarchy aware multi-label predictions (champ)}.

\bibitem[{Wang et~al.(2021)Wang, Hu, Li, and Yu}]{WANG2021106876}
Boyan Wang, Xuegang Hu, Peipei Li, and Philip~S. Yu. 2021.
\newblock \href {https://doi.org/https://doi.org/10.1016/j.knosys.2021.106876} {Cognitive structure learning model for hierarchical multi-label text classification}.
\newblock \emph{Knowledge-Based Systems}, 218:106876.

\bibitem[{Wang et~al.(2022{\natexlab{a}})Wang, Wang, Huang, Sun, and Wang}]{wang2022incorporating}
Zihan Wang, Peiyi Wang, Lianzhe Huang, Xin Sun, and Houfeng Wang. 2022{\natexlab{a}}.
\newblock \href {https://doi.org/10.18653/v1/2022.acl-long.491} {Incorporating hierarchy into text encoder: a contrastive learning approach for hierarchical text classification}.
\newblock In \emph{Proceedings of the 60th Annual Meeting of the Association for Computational Linguistics (Volume 1: Long Papers)}, pages 7109--7119, Dublin, Ireland. Association for Computational Linguistics.

\bibitem[{Wang et~al.(2022{\natexlab{b}})Wang, Wang, Liu, Lin, Cao, Sui, and Wang}]{wang-etal-2022-hpt}
Zihan Wang, Peiyi Wang, Tianyu Liu, Binghuai Lin, Yunbo Cao, Zhifang Sui, and Houfeng Wang. 2022{\natexlab{b}}.
\newblock \href {https://doi.org/10.18653/v1/2022.emnlp-main.246} {{HPT}: Hierarchy-aware prompt tuning for hierarchical text classification}.
\newblock In \emph{Proceedings of the 2022 Conference on Empirical Methods in Natural Language Processing}, pages 3740--3751, Abu Dhabi, United Arab Emirates. Association for Computational Linguistics.

\bibitem[{Wang et~al.(2024)Wang, Wang, and Wang}]{wang2024utilizing}
Zihan Wang, Peiyi Wang, and Houfeng Wang. 2024.
\newblock \href {https://aclanthology.org/2024.lrec-main.1504} {Utilizing local hierarchy with adversarial training for hierarchical text classification}.
\newblock In \emph{Proceedings of the 2024 Joint International Conference on Computational Linguistics, Language Resources and Evaluation (LREC-COLING 2024)}, pages 17326--17336, Torino, Italia. ELRA and ICCL.

\bibitem[{Wehrmann et~al.(2018)Wehrmann, Cerri, and Barros}]{pmlr-v80-wehrmann18a}
Jonatas Wehrmann, Ricardo Cerri, and Rodrigo Barros. 2018.
\newblock \href {https://proceedings.mlr.press/v80/wehrmann18a.html} {Hierarchical multi-label classification networks}.
\newblock In \emph{Proceedings of the 35th International Conference on Machine Learning}, volume~80 of \emph{Proceedings of Machine Learning Research}, pages 5075--5084. PMLR.

\bibitem[{Wolf et~al.(2020)Wolf, Debut, Sanh, Chaumond, Delangue, Moi, Cistac, Rault, Louf, Funtowicz, Davison, Shleifer, von Platen, Ma, Jernite, Plu, Xu, Le~Scao, Gugger, Drame, Lhoest, and Rush}]{wolf-etal-2020-transformers}
Thomas Wolf, Lysandre Debut, Victor Sanh, Julien Chaumond, Clement Delangue, Anthony Moi, Pierric Cistac, Tim Rault, Remi Louf, Morgan Funtowicz, Joe Davison, Sam Shleifer, Patrick von Platen, Clara Ma, Yacine Jernite, Julien Plu, Canwen Xu, Teven Le~Scao, Sylvain Gugger, Mariama Drame, Quentin Lhoest, and Alexander Rush. 2020.
\newblock \href {https://doi.org/10.18653/v1/2020.emnlp-demos.6} {Transformers: State-of-the-art natural language processing}.
\newblock In \emph{Proceedings of the 2020 Conference on Empirical Methods in Natural Language Processing: System Demonstrations}, pages 38--45, Online. Association for Computational Linguistics.

\bibitem[{Wu and Zhou(2016)}]{Wu2016AUV}
Xi-Zhu Wu and Zhi-Hua Zhou. 2016.
\newblock \href {https://api.semanticscholar.org/CorpusID:15209329} {A unified view of multi-label performance measures}.
\newblock In \emph{International Conference on Machine Learning}.

\bibitem[{Yang and Cardie(2013)}]{yang-cardie-2013-joint}
Bishan Yang and Claire Cardie. 2013.
\newblock \href {https://aclanthology.org/P13-1161} {Joint inference for fine-grained opinion extraction}.
\newblock In \emph{Proceedings of the 51st Annual Meeting of the Association for Computational Linguistics (Volume 1: Long Papers)}, pages 1640--1649, Sofia, Bulgaria. Association for Computational Linguistics.

\bibitem[{Yu et~al.(2022)Yu, Shen, and Mao}]{C_F1_v2}
Chao Yu, Yi~Shen, and Yue Mao. 2022.
\newblock \href {https://doi.org/10.1145/3477495.3531765} {Constrained sequence-to-tree generation for hierarchical text classification}.
\newblock In \emph{Proceedings of the 45th International ACM SIGIR Conference on Research and Development in Information Retrieval}, SIGIR '22, page 1865–1869, New York, NY, USA. Association for Computing Machinery.

\bibitem[{Yu et~al.(2023)Yu, He, Basulto, and Pan}]{yu-etal-2023-instances}
Simon Chi~Lok Yu, Jie He, Victor Basulto, and Jeff Pan. 2023.
\newblock \href {https://doi.org/10.18653/v1/2023.findings-emnlp.594} {Instances and labels: Hierarchy-aware joint supervised contrastive learning for hierarchical multi-label text classification}.
\newblock In \emph{Findings of the Association for Computational Linguistics: EMNLP 2023}, pages 8858--8875, Singapore. Association for Computational Linguistics.

\bibitem[{Zangari et~al.(2024)Zangari, Marcuzzo, Rizzo, Giudice, Albarelli, and Gasparetto}]{review}
Alessandro Zangari, Matteo Marcuzzo, Matteo Rizzo, Lorenzo Giudice, Andrea Albarelli, and Andrea Gasparetto. 2024.
\newblock \href {https://doi.org/10.3390/electronics13071199} {Hierarchical text classification and its foundations: A review of current research}.
\newblock \emph{Electronics}, 13(7).

\bibitem[{Zhang et~al.(2022)Zhang, Xu, Soh, and Chen}]{ZHANG2022115922}
Xinyi Zhang, Jiahao Xu, Charlie Soh, and Lihui Chen. 2022.
\newblock \href {https://doi.org/https://doi.org/10.1016/j.eswa.2021.115922} {La-hcn: Label-based attention for hierarchical multi-label text classification neural network}.
\newblock \emph{Expert Systems with Applications}, 187:115922.

\bibitem[{Zhang et~al.(2021)Zhang, Shen, Dong, Wang, and Han}]{zhang2021match}
Yu~Zhang, Zhihong Shen, Yuxiao Dong, Kuansan Wang, and Jiawei Han. 2021.
\newblock Match: Metadata-aware text classification in a large hierarchy.
\newblock In \emph{Proceedings of the Web Conference 2021}, pages 3246--3257.

\bibitem[{Zhou et~al.(2020)Zhou, Ma, Long, Xu, Ding, Zhang, Xie, and Liu}]{zhou-etal-2020-hierarchy}
Jie Zhou, Chunping Ma, Dingkun Long, Guangwei Xu, Ning Ding, Haoyu Zhang, Pengjun Xie, and Gongshen Liu. 2020.
\newblock \href {https://doi.org/10.18653/v1/2020.acl-main.104} {Hierarchy-aware global model for hierarchical text classification}.
\newblock In \emph{Proceedings of the 58th Annual Meeting of the Association for Computational Linguistics}, pages 1106--1117, Online. Association for Computational Linguistics.

\bibitem[{Zhu et~al.(2024)Zhu, Wu, Liu, Hou, Yuan, Li, Pan, and Xu}]{zhu-etal-2024-hill}
He~Zhu, Junran Wu, Ruomei Liu, Yue Hou, Ze~Yuan, Shangzhe Li, Yicheng Pan, and Ke~Xu. 2024.
\newblock \href {https://aclanthology.org/2024.naacl-long.265} {{HILL}: Hierarchy-aware information lossless contrastive learning for hierarchical text classification}.
\newblock In \emph{Proceedings of the 2024 Conference of the North American Chapter of the Association for Computational Linguistics: Human Language Technologies (Volume 1: Long Papers)}, pages 4731--4745, Mexico City, Mexico. Association for Computational Linguistics.

\bibitem[{Zhu et~al.(2023)Zhu, Zhang, Huang, Wu, and Xu}]{zhu2023hitin}
He~Zhu, Chong Zhang, Junjie Huang, Junran Wu, and Ke~Xu. 2023.
\newblock \href {https://doi.org/10.18653/v1/2023.acl-long.432} {{H}i{TIN}: Hierarchy-aware tree isomorphism network for hierarchical text classification}.
\newblock In \emph{Proceedings of the 61st Annual Meeting of the Association for Computational Linguistics (Volume 1: Long Papers)}, pages 7809--7821, Toronto, Canada. Association for Computational Linguistics.

\end{thebibliography}
\newpage
\appendix

\clearpage
\section*{Limitations}


Our work emphasizes fairness and transparency, acknowledging potential limitations within the current framework. However, several key limitations remain. Although we looked for metrics agnostic to the hierarchical context, our core results on metrics and inference are restricted to a specific framework we call \textit{single-path leaf label}. Beyond this framework, theoretical considerations are considerably difficult. Secondly, we demonstrate that the commonly used 0.5 threshold is not optimal for F1-score calculation. Although we address this by considering all possible thresholds for a fair evaluation, each individual instance likely has a unique optimal threshold, which would need further research. 
Finally, our loss functions based on the conditional softmax include the computation of several cascade of conditional probabilities. This means that inaccuracies in probability estimations at higher levels can disproportionately amplify errors at lower levels, potentially compromising overall model performance.
\section{Hierarchical Metrics}
\label{app:hier_metrics_not_used}
In this section, we begin by defining the hierarchical metrics discussed in Section~\ref{sec:hier_f1}. Following that, we evaluate models using these metrics before explaining our decision to exclude them from our main result tables.
\paragraph{Path-constrained F1-score}
As proposed by \cite{C_F1, C_F1_v2}, path-constrained version of multi-label F1-scores (referred as C-metrics in the literature) correspond to enhanced version of standard multi-label F1-scores which better account for correctness. In fact, with these metrics, a node predicted as true will be considered a valid prediction if and only if all its ancestors are also predicted as true. Otherwise, it  will be considered as a false prediction.

\paragraph{Information Hierarchical F1-score.}

Specificity is defined by the level of uncertainty left by predicting a given node: the hF1 metric implicitly assumes an equivalence between specificity and depth. However, predicting a node at a level $i$ can, in some cases, yield a leaf node, representing the highest level of specificity (as it conclusively determines the class prediction with no ambiguity), or it may result in a prediction lacking informativeness when this node has multiple generations of descendants (as it leaves a high level of uncertainty).

 To solve the {specificity} issue,~\citet{NEURIPS2022_727855c3} have proposed an information-based hierarchical F1-score. The intuition is to capture the specificity of a node $y$ through its information content $I$ defined as :  
\begin{equation*}
    I(y) = \log_2|\mathcal{L}| - \log_2|\mathcal{L}\cap\mathcal{D}(y)|
\end{equation*}
where $\mathcal{D}(y)$ is the set of descendants of $y$. For a set $Y= \{y_1, y_2, \dots, y_n\}$, information is defined recursively as follows: 
\begin{align*}
    I(\{y_1, y_2, \dots, y_n\}) = I(&y_1) + I(\{ y_2, \dots, y_n\}) \\&- I(\underset{i=2}{\overset{n}{\cup}}\{\mathrm{lca}(y_1, y_i)\}
\end{align*} where $\mathrm{lca}(y, y')$ denotes the lowest common ancestor of $y$ and $y'$.
The information hierarchical recall $\mathrm{ihR}$ and precision $\mathrm{ihP}$ \matt{are defined similarly to $\mathrm{hR}$ and $\mathrm{hP}$, but using the information content of the sets rather than their cardinality. The ihF1-score is then defined as:}
\begin{equation*}
    \mathrm{ihF1(Y, \hat{Y})} = \frac{2\cdot\mathrm{ihP(Y, \hat{Y})} \cdot \mathrm{ihR(Y, \hat{Y})}}{\mathrm{ihP(Y, \hat{Y})} + \mathrm{ihR(Y, \hat{Y})}}
\end{equation*}

 As the hF1-score, the ihF1-score is agnostic to the label structure~\citep{amigo-delgado-2022-evaluating}, but better accounts for specificity.

\paragraph{Results}
We present comparisons between C-F1-scores and standard F1-scores in Table~\ref{tab:res_C_macro} and Table~\ref{tab:res_C_macro_v2} for all datasets. Our analysis reveals that our top-down loss-based methods yield identical results for both metrics. This outcome is unsurprising since the C-metrics penalize inconsistent predictions, while these methods consistently generate coherent predictions. In contrast, other models show a marginal decrease in \textit{macro} metrics and nearly identical performance in \textit{micro} metrics. These findings lead to two conclusions: firstly, the metrics consistently favor our top-down loss-based methods, and secondly, this preference does not significantly alter the ranking of other models. Consequently, we decided not to include these metrics in the main results tables.\\
 We display in Table~\ref{tab:ihF1_hF1} comparisons between hF1-score AUC and ihF1-score AUC for all datasets. We observe no difference in the rankings of both metrics. Following these observations we decided also not to include this metric within the main table results.

\begin{table*}[!h]        
    \centering
    \resizebox{\textwidth}{!}{
    \begin{tabular}{|l|c|c||c|c||c|c||c|c|}
    \hline
        & \multicolumn{4}{c||}{HWV} & \multicolumn{4}{c|}{WOS}\\ 
        \hline
          \multirow{2}{*}{Method} & \multicolumn{2}{c||}{micro} & \multicolumn{2}{c||}{macro} & \multicolumn{2}{c||}{micro} &\multicolumn{2}{c|}{macro} \\ \cline{2-9}
         & F1 & C-F1 & F1 & C-F1 & F1 & C-F1 & F1 & C-F1 \\ 
        \hline
        BCE   &  85.86$_{\pm 0.15}$ & 85.80$_{\pm 0.17}$& 45.56$_{\pm 0.58}$&44.70$_{\pm 0.58}$ & 87.03$_{\pm 0.05}$ & 87.02$_{\pm 0.06}$ &  81.19$_{\pm 0.12}$ & 81.12$_{\pm 0.15}$\\
        CHAMP  &  87.14$_{\pm 0.15}$& 87.13$_{\pm 0.16}$ & 50.90$_{\pm 0.24}$ & 50.09$_{\pm 0.19}$ & 87.01$_{\pm 0.13}$ & 87.01$_{\pm 0.14}$ & 81.23$_{\pm 0.18}$ &  81.19$_{\pm 0.22}$\\
        \hline
        HBGL  & - & - & - & - &  87.22$_{\pm 0.10}$& 87.21$_{\pm 0.10}$ & 81.86$_{\pm 0.19}$& 81.85$_{\pm 0.19}$ \\
        HGCLR  & 84.92$_{\pm 0.37}$& 84.90$_{\pm 0.37}$ & 44.89$_{\pm 1.38}$ & 44.82$_{\pm 1.31}$& 86.63$_{\pm 0.27}$ & 86.63$_{\pm 0.28}$ & 80.04$_{\pm 0.45}$ & 79.98$_{\pm 0.45}$ \\
        HITIN    & 87.49$_{\pm 0.08}$& 87.47$_{\pm 0.17}$& 51.73$_{\pm 0.42}$ & 51.34$_{\pm 0.60}$ & 87.05$_{\pm 0.10}$ & 87.05$_{\pm 0.10}$ & 81.49$_{\pm 0.06}$ & 81.41$_{\pm 0.08}$\\
        \hline
        Leaf Softmax  & 84.79$_{\pm 0.57}$ & 84.79$_{\pm 0.57}$ & 51.49$_{\pm 0.52}$& 51.49$_{\pm 0.52}$ & 85.91$_{\pm 0.25}$ &85.91$_{\pm 0.25}$ & 80.02$_{\pm 0.29}$ &80.02$_{\pm 0.29}$\\
        Cond. Sigmoid & 87.01$_{\pm 0.19}$ & 87.01$_{\pm 0.19}$& 52.27$_{\pm 0.82}$ & 52.27$_{\pm 0.82}$ & 86.86$_{\pm 0.23}$ &86.86$_{\pm 0.23}$ & 81.07$_{\pm 0.30}$ & 81.07$_{\pm 0.30}$\\
        Cond. Softmax  & 87.49$_{\pm 0.10}$& 87.49$_{\pm 0.10}$& 53.79$_{\pm 0.65}$& 53.79$_{\pm 0.65}$&  86.27$_{\pm 0.17}$ &86.27$_{\pm 0.17}$ & 80.25$_{\pm 0.33}$ &  80.25$_{\pm 0.33}$\\
        Cond. Softmax + LA  (ours) & 87.51$_{\pm 0.07}$&87.51$_{\pm 0.07}$ &54.39$_{\pm 0.58}$ &54.39$_{\pm 0.58}$ & 86.35$_{\pm 0.12}$ &86.35$_{\pm 0.12}$ & 80.11$_{\pm 0.26}$ &80.11$_{\pm 0.26}$ \\
        \hline
    \end{tabular}
    }
    \caption{C-F1-scores vs. standard F1-scores  (and $95\%$ confidence interval) on the test sets of the WOS and HWV datasets for the implemented models}.
    \label{tab:res_C_macro}
\end{table*}

\begin{table*}[!h]
    \centering
    \resizebox{\textwidth}{!}{
    \begin{tabular}{|l|c|c||c|c||c|c||c|c|}
    \hline
        & \multicolumn{4}{c||}{RCV1} & \multicolumn{4}{c|}{BGC}\\ 
        \hline
          \multirow{2}{*}{Method} & \multicolumn{2}{c||}{micro} & \multicolumn{2}{c||}{macro} & \multicolumn{2}{c||}{micro} &\multicolumn{2}{c|}{macro} \\ \cline{2-9}
         & F1 & C-F1 & F1 & C-F1 & F1 & C-F1 & F1 & C-F1 \\ 
        \hline
        BCE  & 86.65$_{\pm 0.30}$ &  86.64$_{\pm 0.30}$ & 66.47$_{\pm 1.49}$ &66.26$_{\pm 1.50}$ & 80.51$_{\pm 0.21}$ & 80.51$_{\pm 0.19}$ & 62.33$_{\pm 1.36}$& 62.03$_{\pm 1.36}$\\
        CHAMP  & 85.93$_{\pm 0.66}$ & 85.88$_{\pm 0.69}$& 62.86$_{\pm 3.64}$ &62.67$_{\pm 3.58}$ &  80.54$_{\pm 0.20}$ & 80.54$_{\pm 0.30}$ & 63.58$_{\pm 0.49}$ & 63.18$_{\pm 0.49}$\\
        \hline
        HBGL  & 87.11$_{\pm 0.12}$ & 87.11$_{\pm 0.12}$& 70.20$_{\pm 0.33}$ & 70.15$_{\pm 0.33}$&  80.19$_{\pm 0.11}$ &80.12$_{\pm 0.09}$ &65.94$_{\pm 0.18}$ & 65.54$_{\pm 0.18}$  \\
        HGCLR  &  86.11$_{\pm 0.26}$ & 86.13$_{\pm 0.26}$& 67.49$_{\pm 0.61}$ & 67.45$_{\pm 0.65}$& 80.16$_{\pm 0.29}$ & 80.15$_{\pm 0.29}$ & 63.58$_{\pm 0.40}$ & 63.38$_{\pm 0.40}$\\
        HITIN  &  85.72$_{\pm 0.60}$ &85.65$_{\pm 0.65}$ & 60.00$_{\pm 5.15}$ & 59.69$_{\pm 5.16}$& 80.36$_{\pm 0.21}$ &80.32$_{\pm 0.20}$ & 61.62$_{\pm 1.47}$ & 61.20$_{\pm 1.57}$\\
        \hline
        Cond. Sigmoid  & 85.77$_{\pm 0.71}$ & 85.77$_{\pm 0.71}$& 63.90$_{\pm 2.45}$ &  63.90$_{\pm 2.45}$ & 80.24$_{\pm 0.46}$&80.24$_{\pm 0.46}$ & 62.65$_{\pm 0.64}$& 62.65$_{\pm 0.64}$\\
        \hline
    \end{tabular}
    }
    \caption{C-F1-scores vs. standard F1-scores  (and $95\%$ confidence interval) on the test sets of the RCV1 and BGC datasets for the implemented models}.
    \label{tab:res_C_macro_v2}
\end{table*}

\begin{table*}[!h]        
    \centering
    \resizebox{\textwidth}{!}{
    \begin{tabular}{|l|c|c||c|c||c|c||c|c|}
    \hline
        & \multicolumn{2}{c||}{HWV} & \multicolumn{2}{c|}{WOS} & \multicolumn{2}{c||}{RCV1} & \multicolumn{2}{c||}{BGC}\\ 
        \hline
          Method & hF1 AUC & ihF1 AUC & hF1 AUC & ihF1 AUC & hF1 AUC & ihF1 AUC & hF1 AUC & ihF1 AUC \\ 
        \hline
        BCE   & 89.23$_{\pm 0.13}$ & 88.33$_{\pm 0.17}$ & 89.18$_{\pm 0.10}$ & 87.16$_{\pm 0.15}$ & 93.66$_{\pm 0.19}$ & 91.88$_{\pm 0.26}$ & 90.26$_{\pm 0.29}$ & 87.93$_{\pm 0.34}$ \\
        CHAMP  & 89.87$_{\pm 0.19}$ & 89.11$_{\pm 0.21}$ & 88.74$_{\pm 0.08}$ & 86.75$_{\pm 0.08}$ & 93.12$_{\pm 0.33}$ & 91.18$_{\pm 0.45}$ & 90.19$_{\pm 0.22}$ & 87.89$_{\pm 0.28}$\\
        \hline
        HBGL  & - & - & 89.00$_{\pm 0.10}$ & 87.16$_{\pm 0.10}$ &  93.35$_{\pm 0.14}$ & 91.69$_{\pm 0.15}$ & 88.08$_{\pm 0.10}$ & 85.63$_{\pm 0.11}$ \\
        HGCLR  & 88.35$_{\pm 0.35}$ & 87.50$_{\pm 0.45}$ & 89.23$_{\pm 0.22}$& 87.24$_{\pm 0.27}$ & 93.27$_{\pm 0.14}$& 91.44$_{\pm 0.16}$ & 89.81$_{\pm 0.17}$& 87.48$_{\pm 0.16}$\\
        HITIN & 90.72$_{\pm 0.16}$ & 90.18$_{\pm 0.15}$ & 88.92$_{\pm 0.04}$ & 87.01$_{\pm 0.04}$ & 93.04$_{\pm 0.24}$ & 91.09$_{\pm 0.36}$ &  90.08$_{\pm 0.16}$ & 87.72$_{\pm 0.23}$\\
        \hline
        Leaf Softmax  & 88.55$_{\pm 0.47}$ & 87.77$_{\pm 0.43}$ & 88.62$_{\pm 0.08}$& 86.71$_{\pm 0.16}$ & - &- & - &- \\
        Cond. Sigmoid & 90.40$_{\pm 0.17}$ &89.63$_{\pm 0.15}$ & 88.78$_{\pm 0.17}$& 86.84$_{\pm 0.16}$& 93.23$_{\pm 0.36}$ & 90.98$_{\pm 0.84}$ & 90.07$_{\pm 0.40}$& 87.75$_{\pm 0.55}$\\
        Cond. Softmax  & 90.94$_{\pm 0.09}$& 90.24$_{\pm 0.13}$ & 88.77$_{\pm 0.07}$ & 86.77$_{\pm 0.07}$  & - &- & - &-\\
        Cond. Softmax + LA  (ours) & 90.97$_{\pm 0.05}$ &90.25$_{\pm 0.09}$ & 88.90$_{\pm 0.10}$ & 86.90$_{\pm 0.13}$ & - &- & - &-\\
        \hline
    \end{tabular}
    }
    \caption{C-F1-scores vs. standard F1-scores  (and $95\%$ confidence interval) on the test sets of the WOS, HWV, RCV1 and BGC datasets for the implemented models}.
    \label{tab:ihF1_hF1}
\end{table*}

\section{Hierarchical-WikiVitals}
\label{sec:hwv}

\subsection{Wikipedia : Vitals articles}

According to Wikipedia, there are almost 7 millions articles on the English Wikipedia, some of which have been selected as vital articles, with 5 levels of "vitality". We chose to work with level 4, consisting in 10k articles from the most-read ones. The sentence "The vital article lists are meant to guide the prioritization of improvements to vital articles and to monitor their quality" is mentioned on Wikipedia. We can therefore reasonably assume that these articles are of high quality, as there are put below a high scrutiny. The vital articles have been categorized by topic into eleven sublists which are : 
\begin{itemize}[itemsep=0pt] 
    \item Technology
    \item Society and social sciences
    \item Arts
    \item Philosophy and religion
    \item Biological and health sciences
    \item Physical sciences
    \item Everyday life
    \item Mathematics
    \item Geography
    \item History
    \item People
\end{itemize}
These elements are themselves categorized into further sublists. This handmade categorization, discussed and peer-reviewed at \url{https://en.wikipedia.org/wiki/Wikipedia_talk:Vital_articles/Level/4}, provides a natural hierarchical categorization of articles which made us construct the HWV dataset. We detail the construction process in the next section.

\subsection{Dataset Construction}

To perform the Wikipedia scrapping, we use a dump of of June 2021. A first gross categorization is performed at \url{https://en.wikipedia.org/wiki/Wikipedia:Vital_articles/Level/4}, resulting in $11$ classes at depth 1. Each category link leads to a page with further sub-categorization. It is very intuitive and we invite the reader to see \url{https://en.wikipedia.org/wiki/Wikipedia:Vital_articles/Level/4/People}, which correspond to this categorization of all article belonging to the \textit{People} category of depth 1. This page shows that among \textit{People}, there are \textit{Entertainers}. Among \textit{Entertainers} there are \textit{Dancers and choreographers} and among them there are those who do \textit{Ballet}. There is no further subcategory and inside \textit{Ballet} we find $7$ Wikipedia articles. One of them is the ballet choreographer \href{https://en.wikipedia.org/wiki/Rudolf_Nureyev}{Rudolf Nureyev}. A natural labellisation for this article is then : $\{$ "People", "Entertainers", "Dancers and choreographers", "Ballet" $\}$. As as representation of the article, we use its abstract. We perform such a methodology for every article and keep track of all the categories encountered to create a hierarchical graph as defined in the 
\hyperref[sec:prob_def]{Problem Formulation Section}.

 Due to categories named identically but being inherently different we need to perform a post-processing step. This involves adding the ancestor category's name to the label for disambiguation. In practice, \mat{the solution simply} corresponds to labeling an article which was obtained through Wikipedia categories $\{$“Maths”, “Algebra” $\}$ as $\{$ “Maths”, “Algebra (Maths)” $\}$ and an article which was obtained through categories $\{$ “Physics”, “Algebra” $\}$ as $\{$ “Physics”, “Algebra (Physics)” $\}$. This can be done without ambiguity as every article belong to a single path of categories and cannot be found in $2$ different leaf category. This thus creates a hierarchy which has a tree structure. \\
This also creates a \textit{Single path leaf label} dataset (see \hyperref[sec:prob_def]{Problem Formulation Section} for full definition), which is very desirable for theoretical considerations. We have indeed leaf label independence which can be useful when computing expectancy without ambiguity, for example. We display in Table~\ref{tab:examples} three full examples extracted from the HWV dataset.\\ 
The dataset (abstracts and their corresponding labels) underwent a preprocessing akin to \citet{zhou-etal-2020-hierarchy} to conform to standard formats and was subsequently divided into train/validation/test splits. It is available within the "data" folder of the code implementation available at \url{https://anonymous.4open.science/r/revisiting_htc-C7EA/README.md}. Corresponding files are \texttt{hwv.taxonomy}, \texttt{hwv\_train.json}, \texttt{hwv\_val.json} and \texttt{hwv\_test.json}.

\subsection{Additional statistics about the dataset}
\begin{figure}[!h]
    \centering
    \includegraphics[width = 0.45\textwidth]{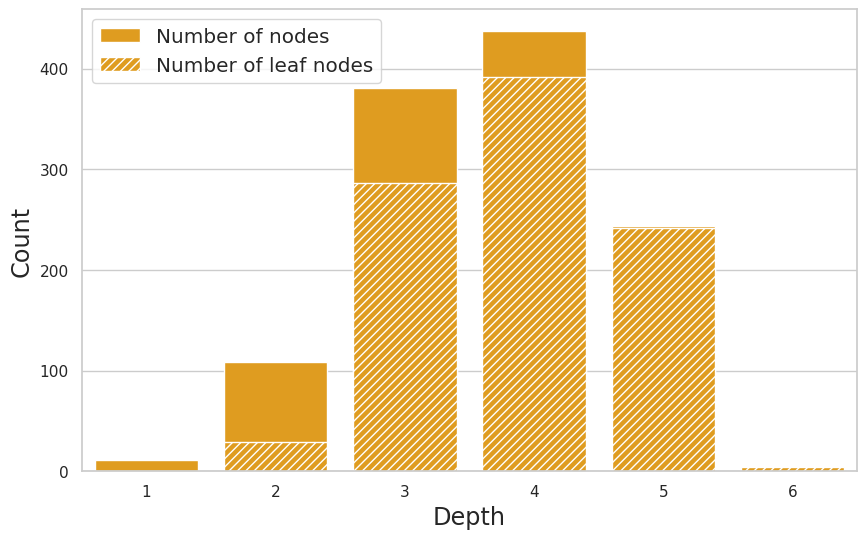}
    \caption{Number of nodes per depth for HWV dataset. Hatched histogram correspond to leaf nodes.}
    \label{fig:depth_nodes}
\end{figure}

The resulting dataset is unique and more challenging that any other HTC dataset. In fact, the hierarchy tree has $1186$ nodes and $953$ leaf nodes, significantly more than any other dataset. We display in Figure~\ref{fig:depth_nodes} the number of nodes and leaf nodes per level. We observe that leaf node depth can vary, ranging from a depth of $2$ to $6$.

\begin{figure}[!h]
    \centering
    \includegraphics[width = 0.45\textwidth]{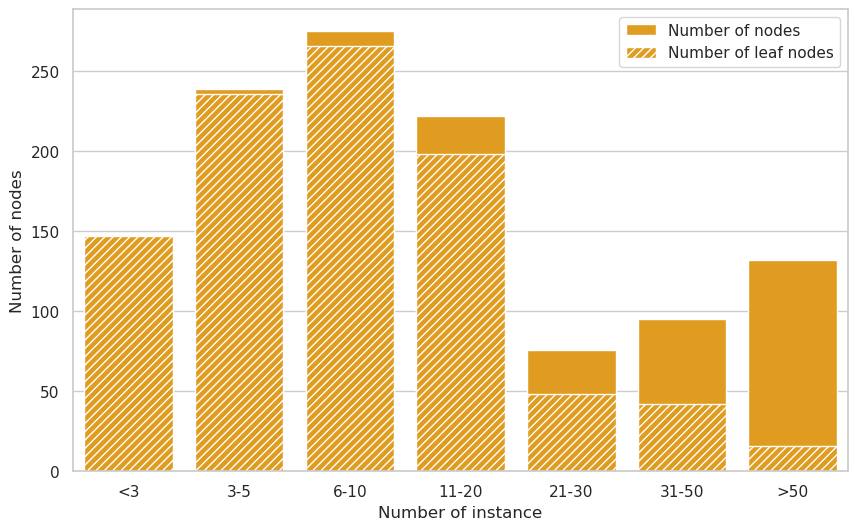}
    \caption{Number of nodes per number of instances in the HWV dataset. Hatched histogram correspond to leaf nodes.}
    \label{fig:number_instances}
\end{figure}

 Another interesting feature of the dataset is the label distribution which is highly imbalanced. We display in Figure~\ref{fig:number_instances} the number of labels per number of instances in that node class. More precisely, it correspond counting the number of instances of each class in the dataset and then create an histogram based on predefined bins of number of instances. We observe that there are a great number of labels that have a number of instances in the dataset smaller than 10. The less training examples we have in the dataset, the harder it is to classify them. We also observe that "fat-tail" effect over leaf nodes, which them also have a great number of classes that have very few examples in the dataset. In addition, we observe that over-represented classes, with $>50$ examples in the dataset are essentially internal nodes of the hierarchy and not leaf nodes.

\section{Proofs}
\subsection{About optimal inference hierarchical metrics}
\label{sec:inf_methodo}
\subsubsection{About $0.5$ thresholding}
\begin{figure}[ht]
    \centering
    \includegraphics[width=0.5\textwidth]{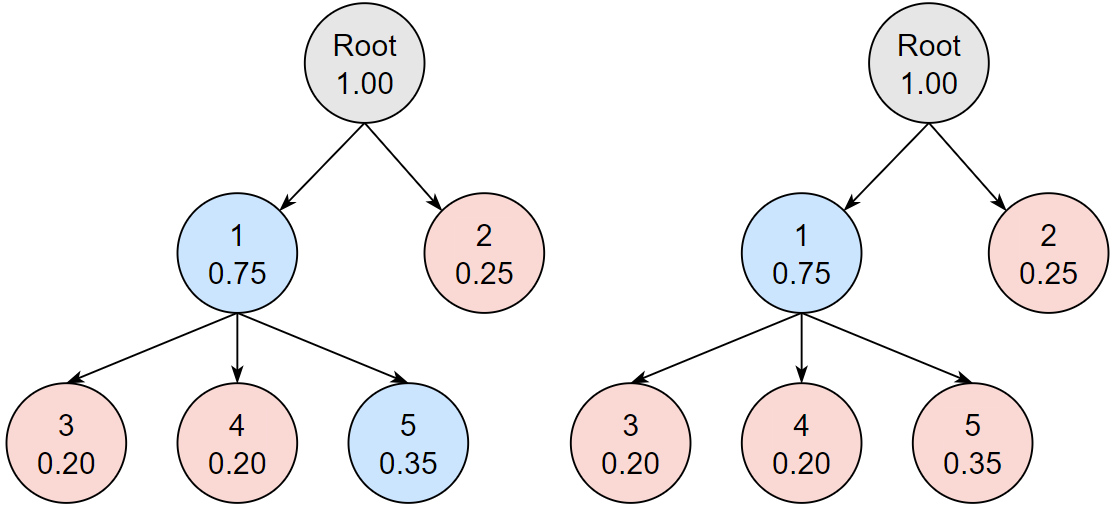}
    \caption{Example of a conditional distribution estimation over a simple hierarchy and corresponding predicted nodes (in blue) for different thresholds(\textit{e.g.} $0.3$ for left case, $0.5$ for right case).}
    \label{fig:ex_g}
\end{figure}


\textbf{hF1-score}
For the left case of Figure~\ref{fig:ex_g} we list all possible events and compute hF1 for each one.
\begin{itemize}
    \item $\mathrm{hF1(\{1,3\}}, \{1, 5\})= \frac{1}{2}$
    \item $\mathrm{hF1(\{1,4\}}, \{1, 5\})= \frac{1}{2}$
    \item $\mathrm{hF1(\{1,5\}}, \{1, 5\})= 1$
    \item $\mathrm{hF1(\{2\}}, \{1\})= 0$
\end{itemize}
Then, 
\begin{align*}
    \mathbb{E}[\mathrm{hF1}(&Y, \{1\})|X=x] \\= &0.2\cdot\frac{1}{2} + 0.2\cdot\frac{1}{2} + 0.35\cdot1 = 0.55
\end{align*}
For the right case of Figure~\ref{fig:ex_g} we list all possible events and compute hF1 for each one.
\begin{itemize}
    \item $\mathrm{hF1(\{1,3\}}, \{1\})= \frac{2}{3}$
    \item $\mathrm{hF1(\{1,4\}}, \{1\})= \frac{2}{3}$
    \item $\mathrm{hF1(\{1,5\}}, \{1\})= \frac{2}{3}$
    \item $\mathrm{hF1(\{2\}}, \{1\})= 0$
\end{itemize}
Then,
\begin{align*}
    &\mathbb{E}[\mathrm{hF1(Y}, \{1\})|X=x] = \\ &0.2\cdot\frac{2}{3} + 0.2\cdot\frac{2}{3} + 0.35\cdot\frac{2}{3} = 0.5
\end{align*}
\begin{equation*}
    \mathbb{E}[\mathrm{hF1(Y}, \{1\})|X=x] < \mathbb{E}[\mathrm{hF1(Y}, \{1, 5\})|X=x] 
\end{equation*}
What we conclude from this simple computation is that it is strictly better to predict node $5$ when aiming at maximizing hF1. We also can conclude that optimal threshold is lower than $0.35$.

\subsubsection{Dependance on $\mathbb{P}(\cdot|x)$ of the optimal threshold}
\label{sec:dep_x}

\begin{figure}[ht]
    \centering
    \includegraphics[width=0.5\textwidth]{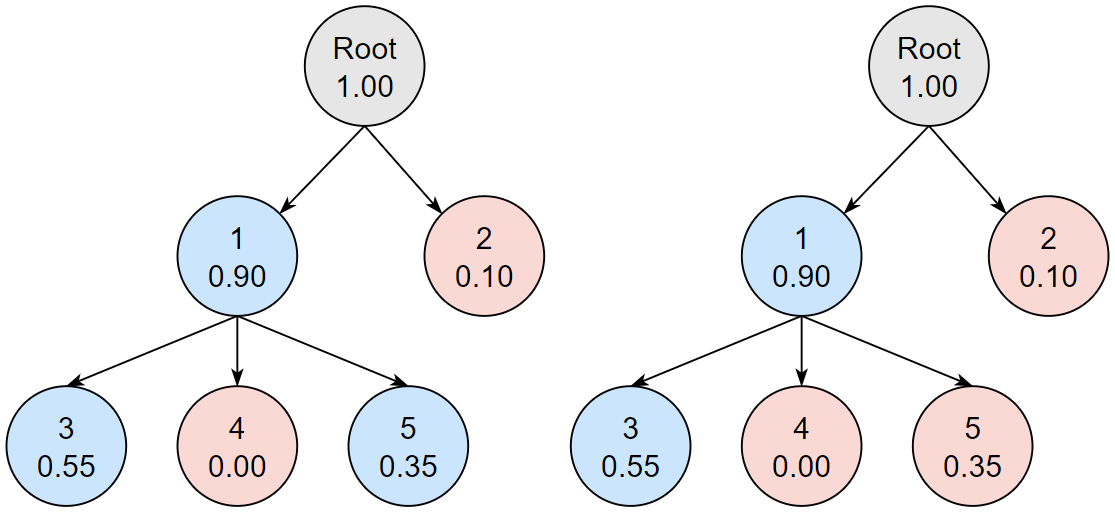}
    \caption{Example of a conditional distribution estimation over a simple hierarchy and corresponding predicted nodes (in blue) for different thresholds(\textit{e.g.} $0.3$ for left case, $0.5$ for right case).}
    \label{fig:ex_g_2}
\end{figure}

 For the left case of Figure~\ref{fig:ex_g_2} we list all possible events and compute hF1 for each one.
\begin{itemize}
    \item $\mathrm{hF1(\{1,3\}}, \{1, 3, 5\})= \frac{4}{5}$
    \item $\mathrm{hF1(\{1,4\}}, \{1, 3, 5\})= \frac{2}{5}$
    \item $\mathrm{hF1(\{1,5\}}, \{1, 3, 5\})= \frac{4}{5}$
    \item $\mathrm{hF1(\{2\}}, \{1\})= 0$
\end{itemize}
Then, 
\begin{align*}
    \mathbb{E}[\mathrm{hF1}(Y, \{1, 3, 5\})|X=x]  &= 0.55 \cdot \frac{4}{5} + 0.0 \cdot \frac{2}{5} \\
     &+ 0.35 \cdot \frac{4}{5} = 0.72
\end{align*}

For the right case of Figure~\ref{fig:ex_g_2} we list all possible events and compute hF1 for each one.
\begin{itemize}
    \item $\mathrm{hF1(\{1,3\}}, \{1, 3\})= 1$
    \item $\mathrm{hF1(\{1,4\}}, \{1, 3\})= \frac{1}{2}$
    \item $\mathrm{hF1(\{1,5\}}, \{1, 3\})= \frac{1}{2}$
    \item $\mathrm{hF1(\{2\}}, \{1\})= 0$
\end{itemize}
Then,
\begin{align*}
    &\mathbb{E}[\mathrm{hF1(Y}, \{1, 3\})|X=x] = \\ &0.55\cdot1 + 0.0\cdot\frac{1}{2} + 0.35\cdot\frac{1}{2} = 0.725
\end{align*}
What we conclude from this simple computation is that it is strictly better to predict node $\{1,3\}$ than  $\{1,3, 5\}$ when aiming at maximizing hF1. We also can conclude that optimal threshold is strictly higher $0.35$ while we proved for the example of Figure~\ref{fig:ex_g} that the optimal threshold was below $0.35$. 

 Both examples shows that the optimal thresholds for each distribution are different and depend on $\mathbb{P}(\cdot|x)$. This naturally leads us to use a \textit{samples} hF1-score, since it makes no sense to compute a F1-score in a \textit{micro} fashion for a given threshold for every $\mathbb{P}(\cdot|x)$.

\subsection{Equivalence between multilabel and hierarchical metrics}
\label{app:equivalence}
Let us consider  $((Y_i, \hat{Y_i}))_{i\in\left[1, N\right]}$
of pairs of targets labels and predicted labels where $$\forall i, ~~ Y_i, \hat{Y_i} \in \{0,1\}^L$$
$L$ is number of different categories.
Let $i \in \left[1, N\right]$ and $j\in \left[1, L\right] $, we denote $Y_i^j$ the $j$-th element of $Y_i$
We define a certain number of metrics below.
\subsubsection{Multi-label F1-score}
We define
\begin{itemize}
    \item The true positives of example $i$ is the set $TP_i = \{j \in \left[1, L\right], ~~(Y_i^j=1)\cap(\hat{Y_i}^j=1) \}$
    \item The true negatives of example $i$ is the set $TN_i = \{j \in \left[1, L\right], ~~(Y_i^j=0)\cap(\hat{Y_i}^j=0)\}$
    \item The false positives of example $i$ is the set $FP_i = \{j \in \left[1, L\right], ~~(Y_i^j=0)\cap(\hat{Y_i}^j=1) \}$
    \item The false negatives of example $i$ is the set $FN_i = \{j \in \left[1, L\right], ~~(Y_i^j=1)\cap(\hat{Y_i}^j=0) \}$
\end{itemize}
\textbf{Micro F1-score}\\
\begin{minipage}{0.45\textwidth}
    \begin{equation*}
    \mathrm{Precision_{micro}} = \frac{\overset{N}{\underset{i=1}{\sum}}\left|TP_i\right|}{\overset{N}{\underset{i=1}{\sum}}\left|TP_i\right| + \left|FP_i\right|}
\end{equation*}
\end{minipage}
\hfill
\begin{minipage}{0.45\textwidth}
    \begin{equation*}
    \mathrm{Recall_{micro}} = \frac{\overset{N}{\underset{i=1}{\sum}}\left|TP_i\right|}{\overset{N}{\underset{i=1}{\sum}}\left|TP_i\right| + \left|FN_i\right|}
\end{equation*}
\end{minipage}

\begin{equation*}
    \mathrm{F_1-score_{micro}} = \frac{2\cdot\mathrm{Precision_{micro}} \cdot \mathrm{Recall_{micro}}}{\mathrm{Precision_{micro}} + \mathrm{Recall_{micro}}}
\end{equation*}
\textbf{Samples F1-score}\\

\begin{equation*}
    \mathrm{Precision_{i}} =  \frac{\left|TP_i\right|}{\left|TP_i\right| + \left|FP_i\right|}
\end{equation*}
    \begin{equation*}
    \mathrm{Recall_{i}} = \frac{\left|TP_i\right|}{\left|TP_i\right| + \left|FN_i\right|}
\end{equation*}
    \begin{equation*}
    \mathrm{F_1-score_{i}} = \frac{2\cdot\mathrm{Precision_{i}} \cdot \mathrm{Recall_{i}}}{\mathrm{Precision_{i}} + \mathrm{Recall_{i}}}
\end{equation*}
\begin{equation*}
    \mathrm{F_1-score_{samples}} = \frac{1}{N}\sum_{i=1}^N \mathrm{F_1-score_{i}}\
\end{equation*}

\subsubsection{Hierarchical F1-score}
\textbf{Micro hF1-score}
\begin{equation*}
    \mathrm{hPrecision_{micro}} = \frac{\overset{N}{\underset{i=1}{\sum}}\left|\hat{Y_i}^{\text{aug}}\cap Y_i\right|}{\overset{N}{\underset{i=1}{\sum}}\left|\hat{Y_i}^{\text{aug}}\right|}
\end{equation*}
\begin{equation*}
    \mathrm{hRecall_{micro}} = \frac{\overset{N}{\underset{i=1}{\sum}}\left|\hat{Y_i}^{\text{aug}}\cap Y_i\right|}{\overset{N}{\underset{i=1}{\sum}}\left|Y_i\right|}
\end{equation*}
\begin{align*}
    &\mathrm{hF_1-score_{micro}} \\&= \frac{2\cdot\mathrm{hPrecision_{micro}} \cdot \mathrm{hRecall_{micro}}}{\mathrm{hPrecision_{micro}} + \mathrm{hRecall_{micro}}}\
\end{align*}
\textbf{Samples hF1-score}\\
\begin{equation*}
    \mathrm{hPrecision_{i}} = \frac{\left|\hat{Y_i}^{\text{aug}}\cap Y_i\right|}{\left|\hat{Y_i}^{\text{aug}}\right|}
\end{equation*}
\begin{equation*}
 \mathrm{hRecall_{i}} = \frac{\left|\hat{Y_i}^{\text{aug}}\cap Y_i\right|}{\left|Y_i\right|}
\end{equation*}
\begin{equation*}
 \mathrm{hF_1-score_{i}} = \frac{2\cdot\mathrm{hPrecision_{i}} \cdot \mathrm{hRecall_{i}}}{\mathrm{hPrecision_{i}} + \mathrm{hRecall_{i}}}
\end{equation*}
\begin{equation*}
    \mathrm{hF_1-score_{samples}} = \frac{1}{N}\sum_{i=1}^N \mathrm{hF_1-score_{i}}\
\end{equation*}

\begin{proposition}
    \label{prop:appendix}
    In \textit{micro} and \textit{samples} settings, if every prediction $\hat{Y}$ is coherent then hF1 and F1 are strictly equal
\end{proposition}

 \textit{Proof.} We recall that we consider here predictions that are  coherent meaning $y\in \hat{Y} \implies \mathcal{A}(y) \subset \hat{Y}.$ In that case $Y_i^{\text{aug}}=Y_i$. In the multi-label framework the micro-precision writes : 
 
\begin{align*}
    \mathrm{Precision_{i}} &= \frac{\left|TP_i\right|}{\left|TP_i\right| + \left|FP_i\right|} \\&= \frac{ \overbrace{\sum_{y\in \hat{Y_i}}\mathbb{1}(y \in Y_i)}^{=\left|\hat{Y_i}\cap Y_i\right|}}{ \underbrace{\sum_{y\in \hat{Y_i}}\underbrace{\mathbb{1}(y \in Y_i) + \mathbb{1}(y \notin Y_i)}_{=1} }_{=\left|\hat{Y_i}\right|}} 
    \\&= \frac{\left|\hat{Y_i}\cap Y_i\right|}{\left|\hat{Y_i}\right|} \\&= \mathrm{hPrecision}_{\text{i}}
\end{align*}
Similarly, 
\begin{align*}
    \mathrm{Recall}_{\text{i}} &= \frac{\left|TP_i\right|}{\left|TP_i\right| +\left|FN_i\right|} \\&= \frac{ \overbrace{\sum_{y\in Y_i}\mathbb{1}(y \in \hat{Y_i})}^{=\left|\hat{Y_i}\cap Y_i\right|}}{ \underbrace{\sum_{y\in Y_i}\underbrace{\mathbb{1}(y \in \hat{Y_i}) + \mathbb{1}(y \notin \hat{Y_i})}_{=1} }_{=\left|Y_i\right|}} 
    \\&= \frac{\left|\hat{Y_i}\cap Y_i\right|}{\left|Y_i\right|} \\&= \mathrm{hRecall}_{\text{i}}
\end{align*}
And naturally, 
\begin{align*}
    \mathrm{hF1-score}_{\text{i}} &=  \frac{2\cdot\mathrm{hPrecision}_{\text{i}}\cdot\mathrm{hRecall}_{\text{i}}}{\mathrm{hPrecision}_{\text{i}} + \mathrm{hRecall}_{\text{i}}} \\&=\frac{2\cdot\mathrm{Precision}_{\text{i}}\cdot\mathrm{Recall}_{\text{i}}}{\mathrm{Precision}_{\text{i}} + \mathrm{Recall}_{\text{i}}} \\&= \mathrm{F1-score}_{\text{i}}
\end{align*}

This computation was performed for \textit{samples} but holds for the \textit{micro} framework. This proves Proposition~\ref{prop:1}
\subsection{Hierarchical logit adjustment}
\label{app:la}
Our motivation is twofold : 
\begin{itemize}
    \item Incorporate prior hierarchy knowledge in our loss
    \item Deal with label imbalance.
\end{itemize}
In \textit{imbalanced} standard classification one typically get rid of standard accuracy metric that can be very high even if witnessing poor results on underrepresented classes. We then want to maximize macro-accuracy. It corresponds to looking for a minimizer of the per-class error rates which writes : 
\begin{equation*}
    \mathrm{BER}(f) = \frac{1}{L}\underset{y\in[L]}{\sum}\mathbb{P}_{x|y}\left(y \notin \underset{y\in[L]}{\mathrm{argmax}}f_{y'}(x)\right)
\end{equation*}
This can be seen as using a \textit{balanced} class probability function $\mathbb{P}^{\text{bal}}(y|x)\propto\frac{1}{L}\mathbb{P}(x|y)$. 

 In our case of hierarchical classification, one typically could want to minimize leaves-balanced error which would lead to minimize 
\begin{equation*}
    \mathrm{BER}(f) = \frac{1}{|\mathcal{L}|}\underset{y\in\mathcal{L}}{\sum}\mathbb{P}_{x|y}\left(y \notin \underset{y\in\mathcal{L}}{\mathrm{argmax}}f_{y'}(x)\right)
\end{equation*}

Let us consider $f^* \in \underset{f:\mathcal{X} \rightarrow \mathbb{R}^{|\mathcal{L}|}}{\mathrm{argmin}}\mathrm{BER}(f)$ the \textit{Bayes-optimal} scorer for this problem. 

Then following \cite{proof1, proof2} we have, 
\begin{equation}
    \label{eq:bay_opt}
    \underset{y\in\mathcal{L}}{\mathrm{argmax}}~f^*_y(x) = \underset{y\in\mathcal{L}}{\mathrm{argmax}}~\mathbb{P}^{\text{bal}}(y|x)
\end{equation}

But, 

\begin{equation*}
    \mathbb{P}^{\text{bal}}(y|x) = \frac{1}{L}\mathbb{P}(x|y) \underbrace{=}_{\text{Bayes formula}} \frac{1}{L}\cdot\frac{\mathbb{P}(y|x)\mathbb{P}(x)}{\mathbb{P}(y)}
\end{equation*}

Then, \eqref{eq:bay_opt} becomes : 

\begin{align}
    \label{eq:bay_opt_v2}
    \underset{y\in\mathcal{L}}{\mathrm{argmax}}f^*_y(x) &= \underset{y\in\mathcal{L}}{\mathrm{argmax}}\frac{1}{|\mathcal{L}|}\cdot\frac{\mathbb{P}(y|x)\mathbb{P}(x)}{\mathbb{P}(y)} \nonumber\\&= \underset{y\in\mathcal{L}}{\mathrm{argmax}} \frac{\mathbb{P}(y|x)}{\mathbb{P}(y)}
\end{align}

Now suppose, as in the conditional softmax framework, that, for a given $y\in\mathcal{Y}$, we have $\mathbb{P}(y|x, \pi(y))\propto\exp{s_y^*(x)}$ for an unknown optimal scorer $s^*:\mathcal{X}\rightarrow\mathbb{R}^{|\mathcal{Y}|}$. 

Then, \eqref{eq:bay_opt_v2} becomes : 
\begin{align}
    &\underset{y\in\mathcal{L}}{\mathrm{argmax}}f^*_y(x) = \underset{y\in\mathcal{L}}{\mathrm{argmax}} \underset{z\in\mathcal{A}(y)}{\prod} \frac{\overbrace{\mathbb{P}(z|x, \pi(z))}^{=\exp(s_z^*(x)})}{\mathbb{P}(z|\pi(z))} \notag \nonumber\\
    &= \underset{y\in\mathcal{L}}{\mathrm{argmax}} \exp\left(\underset{z\in\mathcal{A}(y)}{\sum} s_z^*(x)-\log{\mathbb{P}(z|\pi(z))}\right) \notag \nonumber\\
    &= \underset{y\in\mathcal{L}}{\mathrm{argmax}} \underset{z\in\mathcal{A}(y)}{\sum} s_z^*(x)-\log{\mathbb{P}(z|\pi(z))} \label{eq:logit_adj}
\end{align}

As in \citet{longtail} this suggests training a model to estimate directly $\mathbb{P}^{\text{bal}}$ whose logits are implicitly modified as per \eqref{eq:logit_adj} which would yield the following loss : 
    \begin{equation*}
    l_{\text{CSoLa}}(x, y) =  -\sum_{z \in \mathcal{A}(y)}\log\hat{\mathrm{P}}(z|x, \pi(z))
    \end{equation*}
    Where 
    \begin{equation*}
    \hat{\mathrm{P}}(y|x, \pi(y)) = \frac{e^{s_x^{[y]} + \tau\log\nu(y|\pi(y))}}{\underset{z\in\mathcal{C}(\pi(y))}{\sum}e^{s_x^{[z]} + \tau\log\nu(z|\pi(z))}} 
    \end{equation*}
where $\nu(y|\pi(y))$ is a estimation of  $\mathbb{P}(y|\pi(y))$ and $\tau$ an hyperparameter (which would be optimally $1$).

\subsection{Link between Conditional Softmax and conditional sigmoid}
\label{sec:soft_sig}
\subsubsection{Conditional softmax gradient computation}
\label{sec:cond_softmax_gradient}
\begin{proposition} Let $x \in \mathcal{X}$, $Y\subset\mathcal{Y}$ and $W$ defined as per Equation~\ref{eq:linear_map} then 
\begin{equation*}
    \frac{\partial l_{\mathrm{CSoft}}(x, Y)}{\partial W} = \frac{\partial l_{\mathrm{CSig}}(x, Y)}{\partial W}
\end{equation*}
\end{proposition}
\textit{Proof.} We compute the gradient of the loss with respect to the final weight matrix to understand how parameters of the last layer are updated with the conditional framework. Let first express the loss in terms of the weights of the last layer.
\begin{align*}
    &\mathcal{L}_x = -\sum_{z \in \mathcal{A}(y)}\log\hat{\mathrm{P}}(z|x, \pi(z))\\
    &\mathcal{L}_x= -\sum_{z \in \mathcal{A}(y)}\log(\frac{\exp(W_{[z]}^Th_x + b_{[z]})}{\sum_{j \in \mathcal{C}(\pi(z))}\exp(W_j^Th_x + b_j)})\\
    &\scriptstyle= -\underset{z \in \mathcal{A}(y)}{\sum}\left(W_{[z]}^Th_x + b_{[z]}+ 
\log\left(\underset{{j \in \mathcal{C}(\pi(z))}}{\sum}\exp(W_j^Th_x + b_j)\right)\right)
\end{align*}

Then, we consider the set weights $\mathcal{I}_y = \{ \mathcal{C}(\pi(z)), ~ z \in \mathcal{A}(y)\}$. It correspond to the weights involved in the expression of $\mathcal{L}_x$.

Let $k \in [0, |\mathcal{Y}|-1]$, 
\begin{itemize}
    \item If $k \notin \mathcal{I}_y$ then $$\frac{\partial\mathcal{L}_x}{\partial w_k} = 0$$
    \item If $k \in \mathcal{I}_y$ then 
    \begin{align*}
        \frac{\partial\mathcal{L}_x}{\partial w_k} 
        = -&\mathbb{1}_{k\in\mathcal{A}(y)}h_x \\& + \underbrace{\frac{\exp(w_k^Th_x + b_k)}{\left(\sum_{j \in \mathcal{C}(\pi(k))}\exp(w_j^Th_x + b_j)\right)}}_{\hat{\mathrm{P}}(k|x, \pi(k))} h_x \\
         = -(&\mathbb{1}_{k\in\mathcal{A}(y)} - \hat{\mathrm{P}}(k|x, \pi(k)))h_x
    \end{align*}
\end{itemize}

\subsubsection{Link with Conditional Sigmoid}
\label{sec:contrib_cond_sigmoid}
In Section~\ref{sec:cond_sigmoid}, we introduced the conditional sigmoid, we propose here to provide some justification of the loss employed. (masking BCE introduced in \cite{bertinetto2020})

We recall the definition: 
\begin{equation*}
    \hat{\mathrm{P}}(y|x, \pi(y)) = \frac{1}{1+ \exp{-s_x^{[y]}}} 
\end{equation*}
Where
\begin{equation*}
    s_x = W^Th_x + b ~~ (W\in \mathbb{R}^{d\times |\mathcal{Y}|}, ~b\in \mathbb{R}^{|\mathcal{Y}|})
\end{equation*} 

And then the contribution to the loss of the input text/label $x, y$ is given by Cross-Entropy loss as follows : 
\begin{equation*}
    \scriptstyle
     = -`\underset{z \in \mathcal{A}(y)}{\sum}\left( \log(\hat{\mathrm{P}}(z|x, \pi(z))) + \underset{u \in \mathcal{C}(\pi(z))\backslash\lbrace{z}\rbrace}{\sum} \log(1- \hat{\mathrm{P}}(u|x, \pi(z)))\right)
\end{equation*}

Considering an identical approach as in Section~\ref{sec:cond_softmax_gradient} we show that : 
\begin{itemize}
    \item If $k \notin \mathcal{I}_y$ then $$\frac{\partial\mathcal{L}_x}{\partial w_k} = 0$$
    \item If $k \in \mathcal{I}_y$ then 
        $$\frac{\partial\mathcal{L}_x}{\partial w_k} =  -\left(\mathbb{1}_{k\in\mathcal{A}(y)} - \hat{\mathrm{P}}(k|x, \pi(k))\right)h_x$$
\end{itemize}
Which is exactly the same updates formulas as for the Conditional Softmax. This justifies why we consider such a loss when implementing the conditional sigmoid framework.
\newpage

\begin{table*}[!h]
    \centering
    \resizebox{\textwidth}{!}{
    \begin{tabular}{|p{0.7\linewidth}|p{0.2\linewidth}|}
         \hline
         \textbf{Abstract} & \textbf{Labels}\\
         \hline
         Rudolf Khametovich Nureyev (17 March 1938 - 6 January 1993) was a Soviet-born ballet dancer and choreographer. Nureyev is regarded by some as the greatest male ballet dancer of his generation. Nureyev was born on a Trans-Siberian train near Irkutsk, Siberia, Soviet Union, to a Bashkir-Tatar family. He began his early career with the company that in the Soviet era was called the Kirov Ballet (now called by its original name, the Mariinsky Ballet) in Leningrad. He defected from the Soviet Union to the West in 1961, despite KGB efforts to stop him. This was the first defection of a Soviet artist during the Cold War, and it created an international sensation. He went on to dance with The Royal Ballet in London and from 1983 to 1989 served as director of the Paris Opera Ballet. In addition to his technical prowess, Nureyev was an accomplished choreographer serving as the chief choreographer of the Paris Opera Ballet. He produced his own interpretations of numerous classical works, including Swan Lake, Giselle, and La Bayadère. & People - Entertainers, Dancers and choreographers - Ballet (Dancers and choreographers)\\
         \hline
         Wolfgang Amadeus Mozart (27 January 1756 - 5 December 1791), baptised as Johannes Chrysostomus Wolfgangus Theophilus Mozart, was a prolific and influential composer of the Classical period. Born in Salzburg, in the Holy Roman Empire, Mozart showed prodigious ability from his earliest childhood. Already competent on keyboard and violin, he composed from the age of five and performed before European royalty, embarking on a grand tour. At 17, Mozart was engaged as a musician at the Salzburg court but grew restless and travelled in search of a better position. While visiting Vienna in 1781, he was dismissed from his Salzburg position. He chose to stay in Vienna, where he achieved fame but little financial security. During his final years in Vienna, he composed many of his best-known symphonies, concertos, and operas, and portions of the Requiem, which was largely unfinished at the time of his early death at the age of 35. The circumstances of his death are largely uncertain, and have thus been much mythologized. Despite his early death, his rapid pace of composition resulted in more than 600 works of virtually every genre of his time. Many of these compositions are acknowledged as pinnacles of the symphonic, concertante, chamber, operatic, and choral repertoire. He is considered among the greatest classical composers of all time, and his influence on Western music is profound. Ludwig van Beethoven composed his early works in the shadow of Mozart, and Joseph Haydn wrote: "posterity will not see such a talent again in 100 years". & People - Musicians and composers, Western art music - Composers - Classical (Composers)\\
         \hline
         In probability theory and statistics, a probability distribution is the mathematical function that gives the probabilities of occurrence of different possible outcomes for an experiment. It is a mathematical description of a random phenomenon in terms of its sample space and the probabilities of events (subsets of the sample space).For instance, if X is used to denote the outcome of a coin toss ("the experiment"), then the probability distribution of X would take the value 0.5 (1 in 2 or 1/2) for X = heads, and 0.5 for X = tails (assuming that the coin is fair). Examples of random phenomena include the weather condition in a future date, the height of a randomly selected person, the fraction of male students in a school, the results of a survey to be conducted, etc. & Mathematics - Statistics and probability\\
         \hline
    \end{tabular}
    }
    \caption{Three examples extracted from our newly introduced HWV dataset.}
    \label{tab:examples}
\end{table*}

\end{document}